\documentclass[review]{elsarticle}

\usepackage{hyperref}
\usepackage{picins}
\usepackage{amsmath}
\usepackage{booktabs}
\usepackage{subfig}
\graphicspath{{./pics/}}
\usepackage{amsfonts}       
\journal{Neural Networks}

\begin{document}

\begin{frontmatter}

\title{Distantly Supervised Relation Extraction via Recursive Hierarchy-Interactive Attention and Entity-Order Perception}



\author[addr_1,addr_3]{Ridong Han}
\author[addr_1,addr_2,addr_3]{Tao Peng\corref{mycorrespondingauthor}}
\ead{tpeng@jlu.edu.cn}
\author[addr_4]{Jiayu Han}
\author[addr_1,addr_3]{Hai Cui}

\author[addr_1,addr_2,addr_3]{Lu Liu\corref{mycorrespondingauthor}}
\cortext[mycorrespondingauthor]{Corresponding authors}
\ead{liulu@jlu.edu.cn}

\address[addr_1]{College of Computer Science and Technology, Jilin University, Changchun, Jilin 130012, China}
\address[addr_2]{College of Software, Jilin University, Changchun, Jilin 130012, China}
\address[addr_3]{Key Laboratory of Symbolic Computation and Knowledge Engineering of Ministry of Education, Jilin University, Changchun, Jilin 130012, China}
\address[addr_4]{Department of Linguistics, University of Washington, Seattle, WA 98195, United States}

\begin{abstract}
Wrong-labeling problem and long-tail relations severely affect the performance of distantly supervised relation extraction task. Many studies mitigate the effect of wrong-labeling through selective attention mechanism and handle long-tail relations by introducing relation hierarchies to share knowledge. However, almost all existing studies ignore the fact that, in a sentence, the appearance order of two entities contributes to the understanding of its semantics. Furthermore, they only utilize each relation level of relation hierarchies separately, but do not exploit the heuristic effect between relation levels, i.e., higher-level relations can give useful information to the lower ones. Based on the above, in this paper, we design a novel Recursive Hierarchy-Interactive Attention network (RHIA) to further handle long-tail relations, which models the heuristic effect between relation levels. From the top down, it passes relation-related information layer by layer, which is the most significant difference from existing models, and generates relation-augmented sentence representations for each relation level in a recursive structure. Besides, we introduce a newfangled training objective, called Entity-Order Perception (EOP), to make the sentence encoder retain more entity appearance information.  
Substantial experiments on the popular \emph{New York Times} (NYT) dataset are conducted. Compared to prior baselines, our RHIA-EOP achieves state-of-the-art performance in terms of precision-recall (P-R) curves, AUC, Top-N precision and other evaluation metrics. 
Insightful analysis also demonstrates the necessity and effectiveness of each component of RHIA-EOP.
\end{abstract}


\end{frontmatter}


\section{Introduction}
\label{sec:1}

Various large-scale knowledge bases (KBs), including YAGO \citep{Fabian2007yago}, Freebase \citep{kurt2008freebase} and DBpedia \citep{jens2015dbpedia}, are extremely supportive of many sub-tasks in the field of natural language processing (NLP). However, although existing KBs contain a large number of facts, they are still far from complete compared to the real-world facts, which is infinite. To enrich KBs, many methods have been proposed to automatically extract fact triples from unstructured texts, i.e., relation extraction (RE). 
Among these, supervised approaches are the most commonly used methods and yield relatively high performance. But existing supervised RE systems require massive training data, especially when using neural networks. Furthermore, obtaining high-quality and large-scale training data is very time-consuming and labor-intensive. In this case, distant supervision (DS) is proposed to automatically label training instances by matching KBs to the corpus \citep{mike2009distant}. It assumes that given a pair of entities, all sentences contain these two entities will express the relation between them in KBs. The assumption is too strong and results in some problems.

First, wrong labeling problem is inevitable and has become a bottleneck limiting models' performance by introducing noisy supervision signals. For instance, \textless\textbf{Phil Amicone}, \textbf{Yonkers}\textgreater  expresses the \emph{/people/person/place\_of\_birth} relation in Freebase. So, the sentence ``\emph{Mayor Phil Amicone of Yonkers and the board of Education have supported Mr. Petrone during the controversy.}'' will be automatically labeled as a training sentence, even though it does not express this type of relation. To mitigate the impact of noise labels, multi-instance learning (MIL) \citep{riedel2010modeling, hoffmann2011knowledge} is proposed to identify a relation label for a sentence-bag containing two common entities. Then, many techniques have been introduced into distantly supervised relation extraction (DSRE) task, such as multiple perspectives' attention \citep{lin2016neural, qu2018distant, du2018multi}, soft-labeling \citep{liu2017softlabel}, reinforcement learning \citep{xiao2020joint,yang2020threat}, etc. Among these, the attention mechanism is the most common and successful, so we also extend this technique.

\begin{figure*}[ht]
	\centering
	\includegraphics[width=0.8\linewidth]{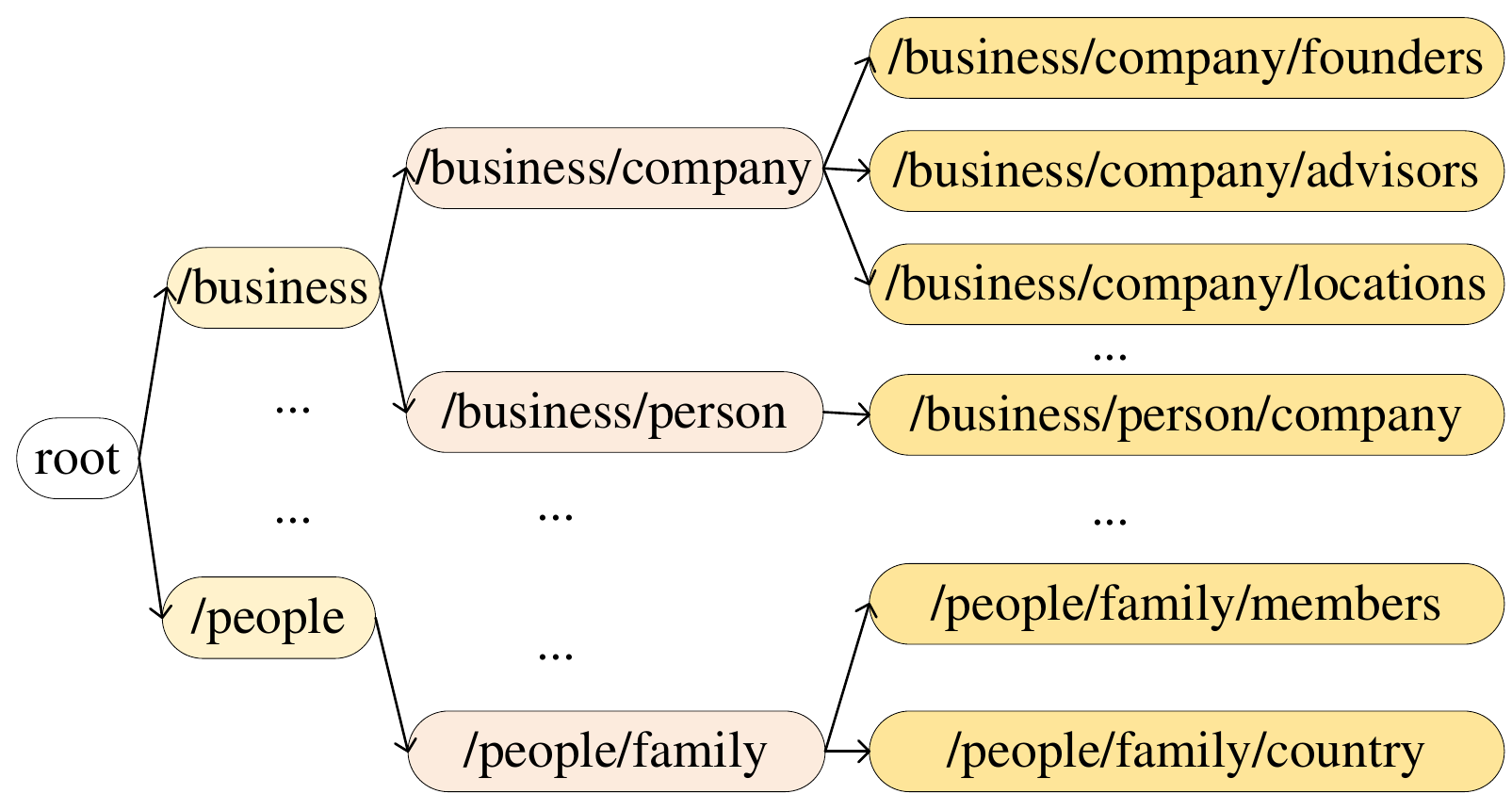}
	\caption{Illustration of the relation hierarchies in Freebase.}
	\label{fig:rel_hier}
\end{figure*} 

Second, although DS method can generate large-scale training data, it can only cover a limited part of real-world relations and causes long-tail problem. For example, in NYT dataset, nearly 77\% of the relations are long-tail.  Here a relation is long-tail if the number of corresponding training instances is less 1000. In this case, data imbalance severely limits the performance of RE systems.
Recently, some approaches naturally share the knowledge from data-rich relations to long-tail ones by leveraging relation hierarchies \citep{han2018hierarchical, zhang2019long, li2020improving, yu2020tohre}. It is based on the observation that although some relations are long-tailed, their ancestor or sibling relations are not. Therefore long-tail relations can benefit from their ancestors or siblings. For example, we select some relations from Freebase and illustrate them in Figure~\ref{fig:rel_hier}. Note that we add a \emph{root} node to facilitate the narrative.
It can be seen that the relation \emph{/business/company/founders} has two ancestor relations (i.e., \emph{/business} and \emph{/business/company}) and several siblings (i.e., \emph{/business/company/advisors}, \emph{/business/company/locations}, etc.). And all relations form a taxonomic structure. With it, 
\citet{han2018hierarchical} propose a hierarchical attention scheme to generate extra bag-level features for each relation level of relation hierarchies separately. Then \citet{zhang2019long} enrich the relation embeddings with TransE \citep{bordes2013transe} and graph convolutional networks \citep{defferrard2016convolutional}. After that, \citet{li2020improving} propose an attention-based method to enhance sentence representations for each relation level independently, while \citet{yu2020tohre} design a Top-Down classification strategy along the relation hierarchies.

However, the above methods all use each relation level of relation hierarchies independently, i.e., although they use multi-grained relation levels to generate extra features or enrich sentence representations, these levels are discrete, independent and do not affect each other in the calculation. In fact, along the hierarchical relation chains, higher-level relations must be instructive to the lower ones. For example, the sentence ``\emph{\textbf{Google} was founded by Larry Page and \textbf{Sergey Brin} on September 4, 1998.}'' expresses the relation \emph{/business/company/founders}. When classifying along the hierarchical relation chains, if it is identified as \emph{/business} at the first level, then at the second level, we can select labels only from the child relations of \emph{/business}. Keep going down until the end of chains, the final label is obtained. During this process, the heuristic effect between relation levels is reflected in layer-by-layer narrowing the scope of the current level's labels based on the classification probability of previous level. It can also be seen as a continuous refinement from coarse to fine granularity.
Distinguishing from existing studies, we aim to implicitly model the heuristic effect through interactions between relation levels, which is one of the most prominent contributions of this paper. 
Considering further, modeling these interactions is somewhat equivalent to exploiting the taxonomic structure of relations to uncover the correlation of relations, which can improve inter-relational discrimination from the side. 

Besides, as we all know, the fact triple $<e_1, r, e_2>$ is not equal with $<e_2, r, e_1>$, therefore the appearance order of two entities is extremely crucial. But during sentence encoding, each feature map generated by CNN/PCNN only retain 1 or 3 maximums along the word sequence through pooling operation. This process ignores the importance of entity order and causes the loss of entity order information, i.e., entity order features are underutilized in the deep learning paradigm.

In this paper, we firstly propose a novel network, named as \textbf{R}ecursive \textbf{H}ierarchy-\textbf{I}nteractive \textbf{A}ttention (RHIA), to fully exploit the relation hierarchies. It assumes that \emph{along the hierarchical relation chains, lower-level relations are influenced by the higher-level ones and current known information}. Based on this, we leverage a recursive structure along the chains to deliver heuristic information about higher-level relations, and obtain the relation-augmented sentence representations.
Then we design a new attention pooling module by using the final hidden state to generate bag-level representations. Besides, to retain more entity-order information in sentence representations, an effective training objective, called Entity-Order Perception (EOP), is introduced. 
Our key contributions are summarized as follows: 
\begin{itemize}
	\item We take the heuristic effect between relation hierarchies into account, then a novel network, called \textbf{RHIA}, is proposed to model this heuristics. It is the first approach in DSRE to uncover the heuristic effect between relation levels in relation hierarchies. 
	\item A newfangled training objective, called \textbf{EOP}, is introduced to improve the expressive ability of  sentence encoder. It enables the sentence representations to retain more entity-order information. 
	\item We conduct substantial evaluation on the widely-used benchmark NYT, and receive state-of-the-art performance in multiple metrics. Insightful analysis also verifies the capability and effectiveness of our RHIA-EOP. The code is released at \url{https://github.com/RidongHan/RHIA-EOP}.
\end{itemize}

\section{Related work}

As an important subtask of natural language processing (NLP), relation extraction (RE) task can be divided into sentence-level extraction \citep{zeng2014relation, geng2020semantic,chen2021aneur}, document-level extraction \citep{xu2021document,huang2021three}, few-shot extraction \citep{yang2021entity}, distantly supervised extraction \citep{qu2018distant,li2020improving,Deng2021anoisy,zhou2021selfselect}, etc. 
Here we 
concentrate on the sentence-level  distantly supervised relation extraction (DSRE) scenario.

For wrong-labeling problem, some people \citep{riedel2010modeling, hoffmann2011knowledge, Surdeanu2012multi} relax the assumption behind DS and develop multi-instance learning (MIL) framework. After that, many techniques have been applied to DSRE. First, \citet{zeng2015distant} improve the sentence encoder with Piecewise Convolutional Neural Networks (PCNN), and identify the instance that most likely expresses the corresponding relation in sentence-bag level. Then, \citet{lin2016neural} design a selective attention among sentences/instances in the sentence level. Inspired by this work, attention models in different aspects are proposed, including word-level attention \citep{qu2018distant}, self-attention \citep{du2018multi}, bag-level attention \citep{yuan2019c2sa, ye2019distant}, feature-level attention \citep{dai2019feature}, segment-level attention \citep{yu2019beyond}, etc. Besides, in order to pick out correct training instances to train the model efficiently, reinforcement learning is introduced into DSRE \citep{xiao2020joint, yang2020threat}. Since the attention mechanism is the most commonly used and successful, we also extend this technique.
 
Relation hierarchies contain taxonomic structure of relations.  
Although some relations are long-tail, their sibling or ancestor relations are not. To handle long-tail relations, existing practices take relation hierarchies into account to share knowledge from higher-level data-sufficient relations to the lower-level long-tail ones, which is intended for long-tail relations to benefit from the training phase of sibling/ancestor relations, i.e., supervised signals come from the relation hierarchies. 
\citet{han2018hierarchical} get advanced performance by introducing a hierarchical attention scheme to derive extra bag-level features. 
\citet{zhang2019long} use TransE \citep{bordes2013transe} and graph convolutional network \citep{defferrard2016convolutional} to obtain relation embeddings, and design a novel attention network along relation hierarchies. After that, \citet{li2020improving} augment the sentence representations with relation embeddings at each level of relation hierarchies to provide more clues to the classifier, while \citet{yu2020tohre} exploit the relation hierarchies to design a top-down classification strategy. Recently, \citet{peng2022ghe-lpc} explore the correlation of relations in the relation hierarchies from both global and local perspectives, aiming to make long-tail relations benefit from their sibling or ancestor relations.
The shortcoming of these models is that the relation levels are discrete, independent and do not affect each other during the calculation, i.e., the heuristic effect between relation levels is ignored. As stated in the Section~\ref{sec:1}, higher-level relations are instructive to lower-level ones when classifying. 
This paper aims to address this flaw, and it is our contribution to differentiating from existing studies. Based on the models of \citet{han2018hierarchical} and \citet{li2020improving}, although our model also generates extra bag-level features and relation-augmented sentence representations, we further design a recursive interaction method to pass the relation-related heuristic information along the relational hierarchical chains.

\section{Our Proposed RE Approach}

\begin{figure*}[ht]
	\centering
	\includegraphics[width=\linewidth]{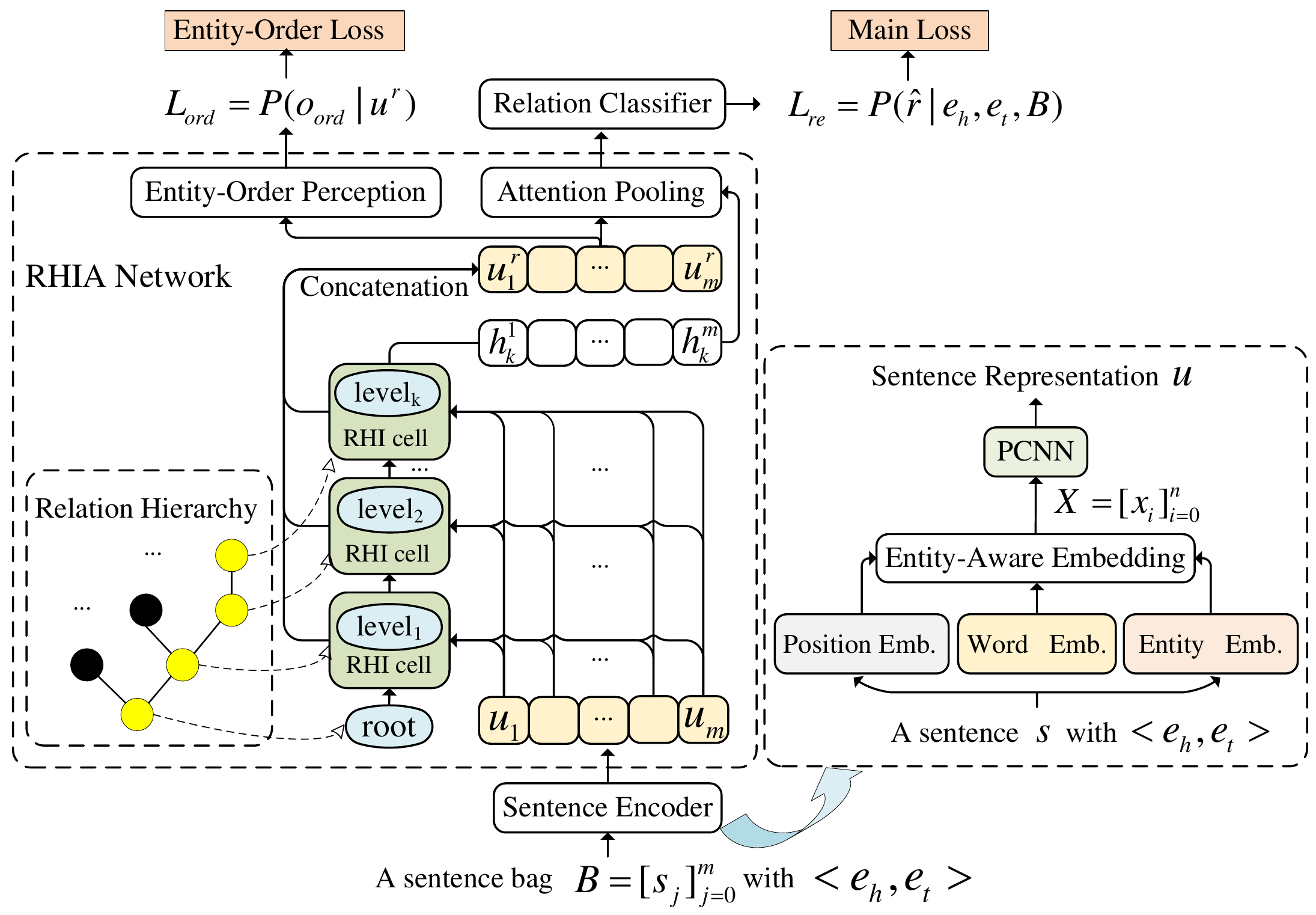}
	\caption{The overview of our proposed RE model, RHIA-EOP.}
	\label{fig:fig4}
\end{figure*}

Our model, RE via Recursive Hierarchy-Interactive Attention and Entity-Order Perception (RHIA-EOP), consists of three cascaded components: 
(1) A sentence encoder based on Entity-Aware Embedding and Piecewise Convolutional Neural Networks (PCNNs). 
(2) A Recursive Hierarchy-Interactive Attention (RHIA) module for generating more valuable bag representations by fully leveraging relation hierarchies. 
(3) A relation classifier module with MultiLayer Perceptron (MLP). The overall model inputs a bag of sentences and outputs the relation label in bag level. Figure~\ref{fig:fig4} shows the overall architecture. 

\subsection{Task Definition}
Given a sentence, the goal of relation extraction (RE) is to identify the relation between a pair of entities in this sentence. 
To facilitate this in distant supervision scenario, 
we split all sentences into multiple entity-pair bags $\{B_1, B_2, ...\}$ . 
Each bag $B_i$ contains some sentences $\{s_1, s_2, ..., s_m\}$ mentioning the same entity pair $<{e_h}_i, {e_t}_i>$. 
Each sentence is a sequence of words, i.e., $s = [w_1, w_2, \dots, w_n]$, and the maximum length is set to $n$. 
Besides, we have a set of pre-defined relation classes $\mathcal{R} = \{r_1, r_2, ...\}$. 
In this case, the goal of DSRE is to distinguish the relation between two given entities based on an entity-pair bag.

\subsection{Sentence Encoder}

In this part, three kinds of features are taken into account, including word embedding \citep{mikolov2013distributed}, position embedding \citep{zeng2014relation} and entity embedding \citep{li2020seg}.
For each sentence $s_j$ in bag $B = \{ s_1, s_2, ..., s_m \}$ \footnote[1]{``sentence'' and ``instance'' are semantically identical.}, we remove the index $j$ for brevity in the following narrative.

\paragraph{Word Embedding}
Each sentence $s=[w_1, w_2, ..., w_n]$ is translated into low-dimensional embeddings, i.e., $V = [v_1, v_2, ..., v_n] \in \mathbb{R}^{d_w \times n}$, where $d_w$ denotes the dimension of word embedding. 

\paragraph{Position Embedding}
Relative position information is very vital to RE task \citep{zeng2014relation}, which is defined as the combination of relative distances from each word to entity $e_h$ and entity $e_t$. Take the sentence ``\emph{It showed that \textbf{Sergey Brin}, a co-founder of \textbf{Google}, not his partner, Larry Page, who is speaking at the conference.}'' as an example, the relative distance from \emph{co-founder} to entity $e_h$ (\emph{Google}) and entity $e_t$ (\emph{Sergey Brin}) are -2 and 3, respectively. Then, two low-dimensional vectors, $p_i^{e_h}$ and $p_i^{e_t} \in \mathbb{R}^{d_p}$ are converted from these two distances. In this way, we can define position-aware embeddings as $X^p = [x_1^p, x_2^p, ..., x_n^p] \in \mathbb{R}^{(d_w + 2d_p) \times n}$, where $x_i^p = [v_i; p_i^{e_h}; p_i^{e_t}]$, $i \in [1,2,...,n]$, ``$;$'' is vector concatenation operation. 

\paragraph{Entity embedding}
The sequence of entity embeddings is represented as $X^e = [x_1^e, x_2^e, ..., x_n^e] \in \mathbb{R}^{3d_w \times n}$, where $x_i^e = [v_i; v_{e_h}; v_{e_t}]$, $i \in [1,2,...,n]$, $v_{e_h}$ and $v_{e_t}$ are the embeddings of two entities. We employ the same way as \citet{li2020improving} to obtain the embeddings of entities. Each entity is one entry in the vocabulary of word embedding even if it is usually composed of multiple words. To achieve this, if an entity consists of multiple words, all words are connected with ``\_'' to denote it. 

\paragraph{Entity-Aware Embedding}
To make better use of the above features, a position-wise gate \citep{li2020seg} is used to integrate them, i.e., 

\begin{align}
  &A^e = sigmoid(\lambda \cdot (W^{e}X^e + b^{e})) \label{Eq-1}\\
  &\tilde{X^p}= tanh(W^{p}X^p + b^{p}) \label{Eq-2}\\
  &X = A^e \circ X^e + (1-A^e) \circ \tilde{X^p} \label{Eq-3}
\end{align}
where ``$\circ$'' indicates the element-wise product, $W^{e} \in \mathbb{R}^{d_x \times 3d_w}$, $W^{p} \in \mathbb{R}^{d_x \times (d_w+2d_p)}$, $\lambda$ is a trade-off weight. And $X = [x_1, x_2, ..., x_n] \in \mathbb{R}^{d_x \times n}$ is the resulting representation of words for encoding.

\paragraph{Piecewise Convolutional Neural Networks}

We select the Piecewise Convolutional Neural Networks (PCNNs) as the sentence encoder \citep{zeng2015distant} because of its high performance and efficiency. 
Given the input representation $X$, PCNN applies a kernel of window size $\omega$ to slide over $X$, and output feature representation $f$, where $f \in \mathbb{R}^{d_c \times n}$ and $d_c$ is the number of filters.
After that, the feature $f$ is firstly divided into three segments $\{f^{(1)}, f^{(2)}, f^{(3)}\}$ based on the position of two entities. And then, the max-pooling operation is employed on each segment, respectively. The results are concatenated as the final sentence representation $u$:

\begin{align}
	u = [max(f^{(1)}); max(f^{(2)}); max(f^{(3)})] \label{Eq-4}
\end{align}
where $u \in \mathbb{R}^{d_f}, d_f =3d_c$.

\subsection{Recursive Hierarchy-Interactive Attention Network}
 
\begin{figure}[!ht]
	\centering
	\includegraphics[width=0.75\linewidth]{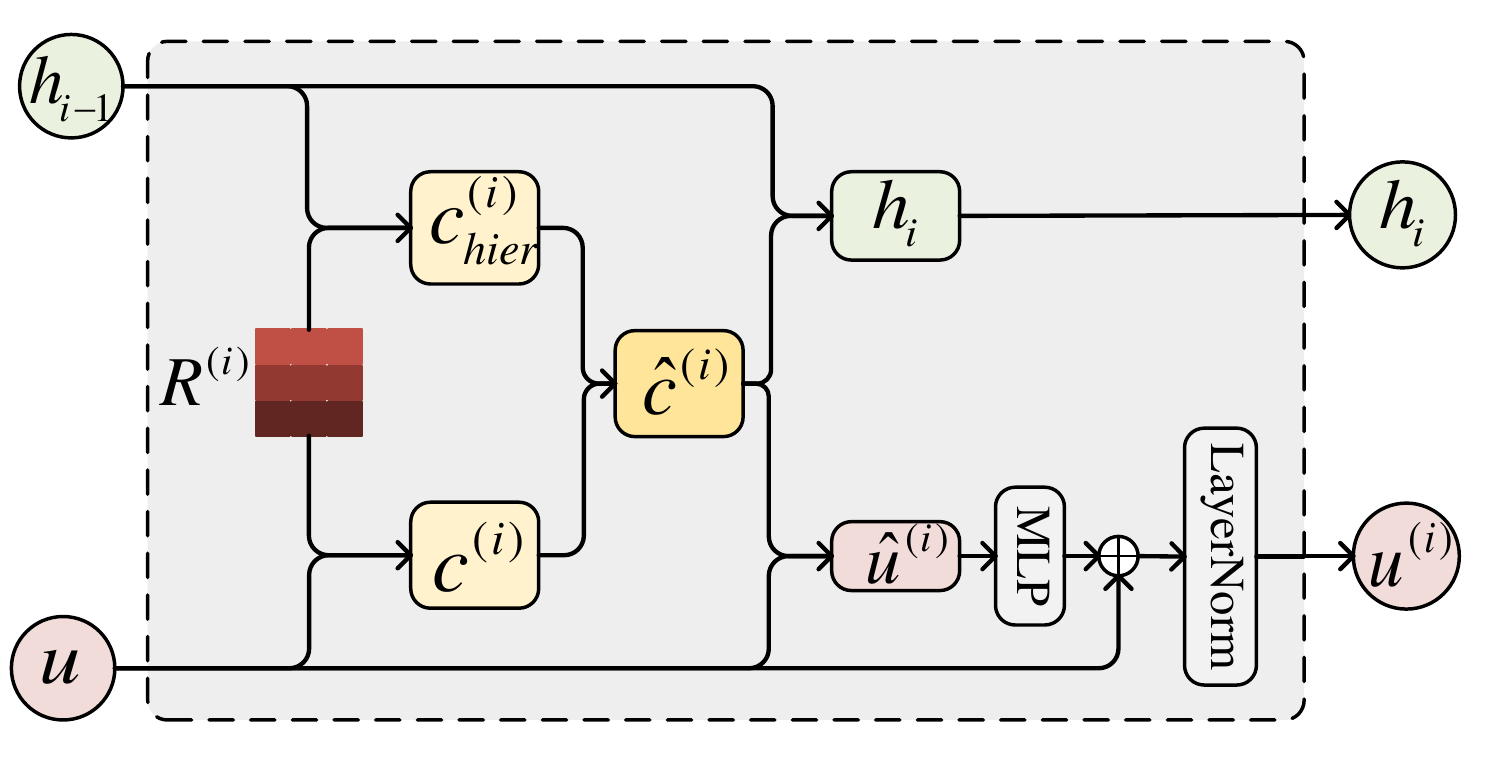}
	\caption{The process of recursive computing unit, RHI cell.}
	\label{fig:fig5}
\end{figure}

To fully exploit the taxonomic structure of relations and relation embeddings, an attention network with a recursive structure along the relation hierarchies, called RHIA, is proposed. RHIA consists of several RHI cells. Each cell completes the calculation for one relation level. See Figure~\ref{fig:fig5} for the details of the RHI cell.

For a bag $B = \{s_1, s_2, ..., s_m \}$, the sentence representations $U = \{u_1, u_2, ..., u_m\}$ are derived via PCNNs encoder. 
For each relation $r \in \mathcal{R}$, its hierarchical chain of relations $\{r^0, r^1, ..., r^{k} \}$ can be generated, where $r^0$ is the \emph{root} relation node and $r^{k}$ is identical to $r$. Connected by the \emph{root}, all chains compose the relation hierarchies, a kind of tree-like structure. Then, for the $i$-th level of relation hierarchies ($i \in [1, 2, ..., k]$), we define a learnable relation embedding matrix $R^{(i)} \in \mathbb{R}^{d_f \times N^i}$, where $N^i$ denotes the number of relations at level $i$. 

For a sentence representation $u \in U$ \footnote[2]{For convenience, we remove indices in the remaining parts.}, we aim to augment it with the above relation embeddings for each level, and all levels' augmented representations are concatenated as the relation-augmented sentence representation $u^{r}$. Then an attention-pooling module is employed to generate the bag-level representation. 
For $i$-th level, it is assumed that \emph{the augmented representation of $i$-th level is determined by the input sentence representation $u$ and the heuristic information about relations $h_{i-1}$ from the higher/previous level }. 

Based on this, we can build the network in a recursive structure.
In details, the sentence-to-relation (sent2rel) attention \citep{li2020improving} is employed firstly. The sentence representation $u$ and the heuristic information $h_{i-1}$ are used as the query to calculate attention scores by dot product with the relation embedding matrix $R^{(i)}$, respectively,

\begin{align}
	& \alpha ^{(i)} = softmax(u^T R^{(i)}) \label{Eq-5}\\
	&c^{(i)} = R^{(i)} \alpha^{(i)} \label{Eq-6}\\
	& \alpha _{hier} ^{(i)} = softmax({h_{i-1}}^T R^{(i)}) \label{Eq-7}\\
	&c_{hier}^{(i)} = R^{(i)} \alpha_{hier}^{(i)} \label{Eq-8}
\end{align}
where $h_{i-1} \in \mathbb{R}^{d_f}$, $softmax(\cdot)$ is an activation function for the last dimension, $c^{(i)}$ and $c^{(i)}_{hier}$ are the relation-aware information. 

Since the importance of $u$ and $h_{i-1}$ are different, then, we leverage an element-wise gate mechanism to integrate the relation-aware information $c^{(i)}$ and $c_{hier}^{(i)}$:

\begin{align}
	& \beta_1^{(i)} = sigmoid(W^{g1}[u; h_{i-1}] + b^{g1}) \label{Eq-9}\\
	& \hat{c}^{(i)} = \beta_1^{(i)} \circ c^{(i)} + (1-\beta_1^{(i)}) \circ c_{hier}^{(i)} \label{Eq-10}
\end{align}
where $W^{g1} \in \mathbb{R}^{2d_f \times d_f}$, $\hat{c}^{(i)}$ is the resulting relation-aware representation corresponding to $i$-th level. 

After that, to prevent the information loss of the original representation $u$, we leverage an element-wise gate to inject the information of $\hat{c}^{(i)}$ into $u$. 
In this process, residual connection \citep{he2016residual} and layer normalization \citep{ba2016layer} are also applied.
Then, the augmented representation $u^{(i)}$ at level $i$ is generated,

\begin{align}
	&\beta_2^{(i)} = sigmoid(W^{g2}[u; \hat{c}^{(i)}] + b^{g2}) \label{Eq-11}\\
	&\hat{u}^{(i)} = \beta_2^{(i)} \circ u + (1-\beta_2^{(i)}) \circ \hat{c}^{(i)} \label{Eq-12}\\
	&u^{(i)} = LayerNorm(u + MLP(\hat{u}^{(i)})) \label{Eq-13}
\end{align}
where $W^{g2} \in \mathbb{R}^{2d_f \times d_f}$, $MLP(\cdot)$ is a multi-layer perceptron that aims to increase nonlinearity.

Finally, we update $h_{i-1}$ to obtain the current heuristic information about relations $h_{i}$ for the next level's calculation, which is achieved by merging relation-aware information $\hat{c}^{(i)}$ into $h_{i-1}$ with an element-wise gate,

\begin{align}
	&\beta_3^{(i)} = sigmoid(W^{g3}[h_{i-1}; \hat{c}^{(i)}] + b^{g3}) \label{Eq-14}\\
	&h_{i} = \beta_3^{(i)} \circ h_{i-1} + (1-\beta_3^{(i)}) \circ \hat{c}^{(i)} \label{Eq-15}
\end{align}
where $W^{g3} \in \mathbb{R}^{2d_f \times d_f}$.

During the calculation, $h_0$ is randomly initialized. 
Along the relation hierarchies, all levels' augmented representations are generated, i.e., $\{u^{(1)}, u^{(2)}, ..., u^{(k)}\}$. Then the relation-augmented sentence representation $u^r$ is generated by concatenation operation,

\begin{align}
	u^{r} = [u^{(1)}; u^{(2)};  ...; u^{(k)}] \label{Eq-16}
\end{align}

For the bag $B$, all relation-augmented sentence representations can be denoted as $B^{r} = [u_1^{r}, u_2^{r}, ..., u_m^{r}]$.
Next, to alleviate wrong labeling problem, we use the attention-pooling \citep{lin2017self, shen2018disan} to select the correctly labelled sentences from the bag in order to facilitate the generation of an accurate bag-level representation. The attention score for each sentence is calculated from its original representation $u$ and its final hidden state $h_k$ (i.e., the $k$-th level's heuristic information about relations). Then a weighted sum over the bag is employed,

\begin{align}
	b = B^{r} softmax(W_{att}^T [U; H]) \in \mathbb{R}^{kd_f} \label{Eq-17}
\end{align}
where $H$ is the matrix consisting of the final hidden state $h_{k}$ of all sentences in $B$.

Finally, a softmax classifier based on the MultiLayer Perceptron is used to classify the bag representation $b$, 

\begin{align}
	o_b = P(\hat{r}|e_h,e_t,B) = softmax(MLP(b)) \label{Eq-18}
\end{align}
where $o_b \in \mathbb{R}^{|\mathcal{R}|}$, $|\mathcal{R}|$ is the number of pre-defined relations.

\subsection{Entity-Order Perception and Training Objectives}

To retain more entity-order information, we design a classification sub-task in the multi-task paradigm. Specifically, a MultiLayer Perceptron is employed to bicategorize relation-augmented representations $u^{r}$, i.e., whether entity $e_h$ appears before entity $e_t$ or not. That is,

\begin{align}
	P(o_{ord}|u^{r}) = softmax(MLP(u^{r})) \label{Eq-19}
\end{align}

To optimize our model, three objectives are introduced:
1) The main objective is bag-level classification and is defined as minimizing cross-entropy loss, 

\begin{equation}
  L_{re} = - \frac{1}{|D|} \sum\limits_{B \in D} {\log{P(\hat{r}|e_h,e_t,B)}} \label{Eq-20}
\end{equation}
where $D$ is the train set consisting of sentence bags. 
2) The hierarchical auxiliary objective is designed to guide RHIA module in choosing appropriate relation embeddings to augment each sentence representation. That is,

\begin{align}
  L_{hier} = - \frac{1}{|D| \times |B| \times k} \sum\limits_{B \in D} {\sum\limits_{s \in B} {\sum\limits_{l=1}^k {\log{\alpha ^{(l)}_{[r^l]}}}}} \label{Eq-21}
\end{align}
where $[\cdot]$ denotes indexing operation.	
3) The entity-order perception objective is introduced to take entity-order information into account:

\begin{align}
  L_{ord} = - \frac{1}{|D| \times |B|} \sum\limits_{B \in D} {\sum\limits_{s \in B} {\log{P(o_{ord}|u^{r})}}}
\end{align}

Eventually, these three objectives are integrated into a whole. The final loss function can be represented as:

\begin{align}
	L = L_{re} + \mu L_{hier} + \xi L_{ord} + \delta ||\theta||_2^2
\end{align}
where $\mu$, $\xi$ and $\delta$ are weight scores of different losses. $||\theta||_2^2$ is the $L_2$ regularizer. 

\section{Experiments and Results}

Since this paper focuses on the sentence-level distantly supervised relation extraction task, we choose the \emph{New York Times} (NYT) dataset \citep{riedel2010modeling} for evaluation\footnote[3]{http://iesl.cs.umass.edu/riedel/ecml/}, which is the only widely-adopted distantly supervised relation extraction benchmark and freely available\footnote[4]{https://github.com/thunlp/HNRE/tree/master/raw\_data}. 
The wrong labeling problem and long-tail problem are extremely serious on it, which is why we choose it. As for other available DS datasets, GIDS \citep{jat2018improving} has no the long-tail phenomenon; NYT-H \citep{zhu2020towards} is subset variants of NYT, and its test set contains only 9955 sentences, which is so small that the results of our model and baselines are all very high and not comparable; DocRED \citep{yao2019docred} is designed for document-level relation extraction that require inter-sentence reasoning capability.  

The NYT dataset was generated by aligning the corpus with Freebase, and has 53 relations. Among these relation types, there is a special \emph{NA} class which indicates that no relation exists between two entities. Its train set consists of the corpus from the years 2005-2006, while the test set consists of the rest corpus from year 2007. 
For its statistics, the number of sentence and entity\_pair of train set are 570088 and 293162, respectively, while the values for the test set are 172448 and 96678. 
 
Our RHIA-EOP is programmed using the Pytorch framework and trained on the GeForce GTX 1080 Ti. During the training phase, it takes about 1.5 hours to execute 15-20 epochs before convergence. Besides, following previous work \citep{lin2016neural}, three-fold cross-validation on the train set is used for hyper-parameter selection and the held-out evaluation is applied to conduct the experiments. 
The evaluation metrics used in this paper include \textbf{Top-N precision}, \textbf{the precision-recall curve}, \textbf{AUC}, \textbf{Max\_F1} and \textbf{Hits@K}.

\subsection{Experimental Settings}

For the initialization of word embeddings, we employ the pre-trained embeddings from \citet{lin2016neural} \footnote[5]{https://github.com/thunlp/OpenNRE}. The relation embedding matrices are initialized randomly as \citet{li2020improving}. Besides, the dropout strategy \citep{srivastava2014dropout} is applied to the bag representations to prevent overfitting. During training phase, for optimization, we employ mini-batch SGD \citep{cotter2011minibatch} with the initial learning rate $\gamma$. 

\begin{table}[!ht]
  \centering
  \caption{Parameters settings.}
  \label{tab:table1}
  \resizebox{0.66\linewidth}{!}{
	\begin{tabular}{cll}
		\toprule
		Parameter     & Value \\
		\midrule
		word/entity embedding dimension $d_w$ & 50     \\
		position embedding dimension $d_p$   & 5      \\
		maximum length of sentences $n$     & 120  \\
		entity-aware smoothness $\lambda$  & 0.05 \\
		Entity-Aware Embedding dimension $d_x$ & 150 \\
		kernel size $\omega$     & 3  \\
		hidden dimension of PCNN $d_c$     & 230  \\
		learning rate $\gamma$     & 0.1 \\
		dropout rate      & 0.5  \\
		number of relation levels $k$ & 3 \\ 
		$L_2$ regularization coefficient $\eta$     & 1e-5  \\
		weights of loss function $\mu, \xi, \delta$   & 1, 1, 1 \\
		\bottomrule
  \end{tabular}}
\end{table}

The detailed parameter settings are shown in Table~\ref{tab:table1}. For Entity-Aware Embedding module and PCNN module, the parameters keep consistent with those in previous work \citep{li2020improving}, i.e., $d_w, d_p, n, \lambda, \omega, d_c$. The learning rate and dropout rate are also the same as \citet{ han2018hierarchical}. 
For parameters of three objective functions, all are consistent with \citet{li2020improving} except for the weight of $L_{ord}$.
To better exploit Entity-Order Perception, we choose the weight of $L_{ord}$ (i.e., $\xi$) from the list $[0.4, 0.5, 0.6, 0.7, 0.8, 0.9, 1.0, 1.5, 2.0, 3.0]$. 
Here we use three-fold cross-validation on the train set.
Each value is trained three times since cross-validation, and the best value (i.e., 1) is determined according to the average of AUC. 

For the following experiments, we apply the held-out evaluation to evaluate our model. The entire train set is used for the training phase, while the test set is used for evaluation. The results in Section~\ref{subsec:sta} are only intended to illustrate the stability of our model, and are not relevant to cross-validation for hyper-parameter selection.

\subsection{Baselines}
We compare RHIA-EOP with competitive previous baselines that are summarized as follows: 

\begin{itemize}
  \item \textbf{PCNN\_ATT}: \ \citet{lin2016neural} propose a selective attention among training sentences to mitigate wrong labeling problem, which is the most classical approach in DSRE task.
  \item \textbf{HNRE}: \ \citet{han2018hierarchical} design a hierarchical attention network to enrich bag-level representations, which is the first hierarchical relation extraction baseline.
  \item \textbf{ToHRE}: \ \citet{yu2020tohre} introduce a top-down classification strategy and a method to enhance the bag representation in different relation levels.
  \item \textbf{CoRA}: \ \citet{li2020improving} 
  enhance sentence representations in a collaborative way across all relation levels.
  \item \textbf{HNRE w/ Ent. Emb.}: The model of \citet{han2018hierarchical} directly introduces the Entity-Aware Embedding layer in Eqs.~\ref{Eq-1}-\ref{Eq-3}. 
  \item \textbf{HNRE w/ Aux. Obj.}: The model of \citet{han2018hierarchical} directly introduces the hierarchical auxiliary objective in Eq.~\ref{Eq-21}. 
  \item \textbf{HNRE w/ Ent. Emb. w/ Aux. Obj.}: The model of \citet{han2018hierarchical} introduces both the Entity-Aware Embedding layer in Eqs.~\ref{Eq-1}-\ref{Eq-3} and the hierarchical auxiliary objective in Eq.~\ref{Eq-21}. 
\end{itemize}

\subsection{Model Comparison Results}

\begin{figure*}[ht]
  \centering
  \subfloat[model comparison.]{
    \begin{minipage}{0.51\linewidth}
      \centering
      \includegraphics[width=0.99\linewidth]{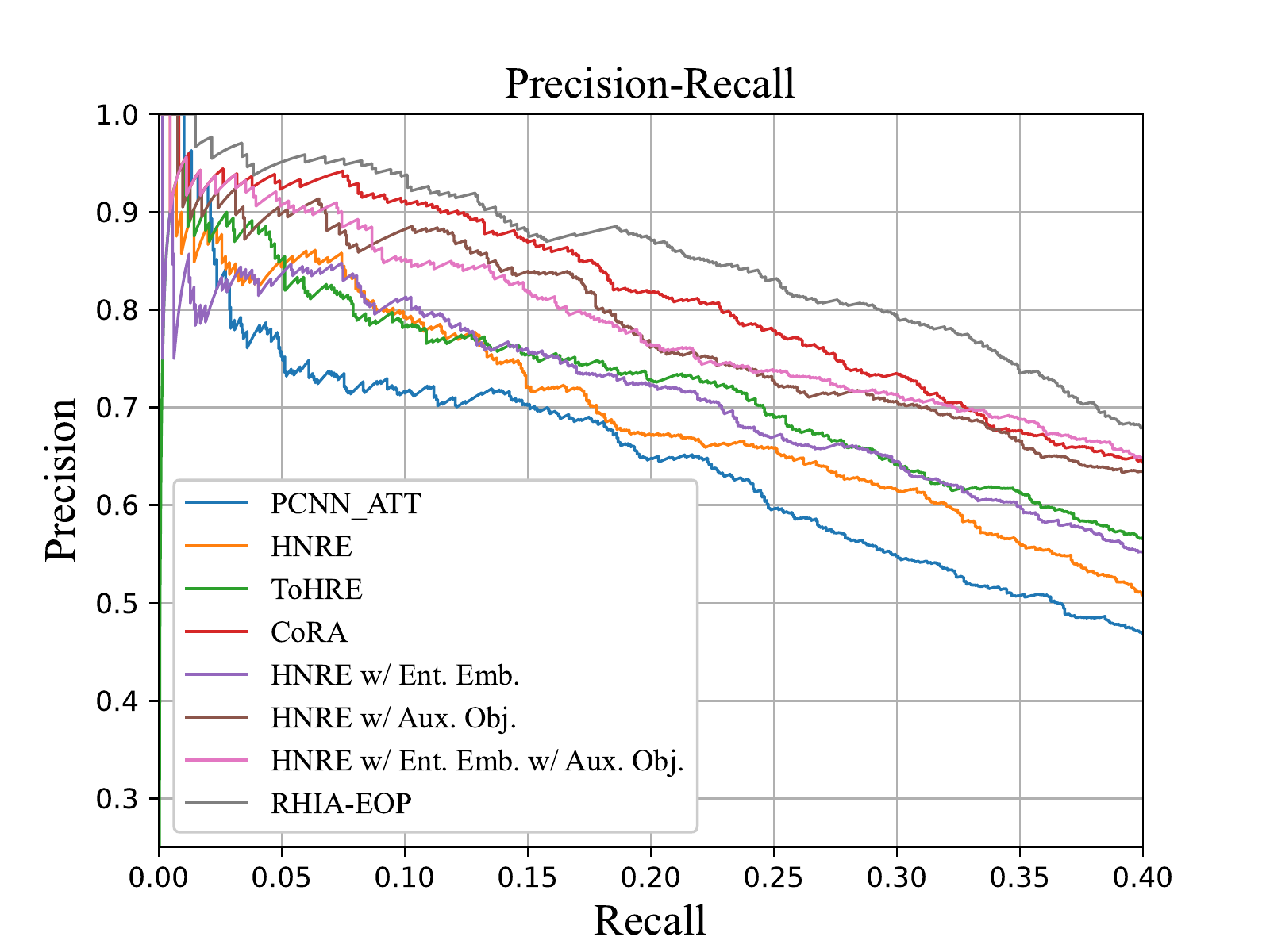}
    \end{minipage}
  }
  \subfloat[model comparison.]{
    \begin{minipage}{0.51\linewidth}
      \centering
      \includegraphics[width=0.99\linewidth]{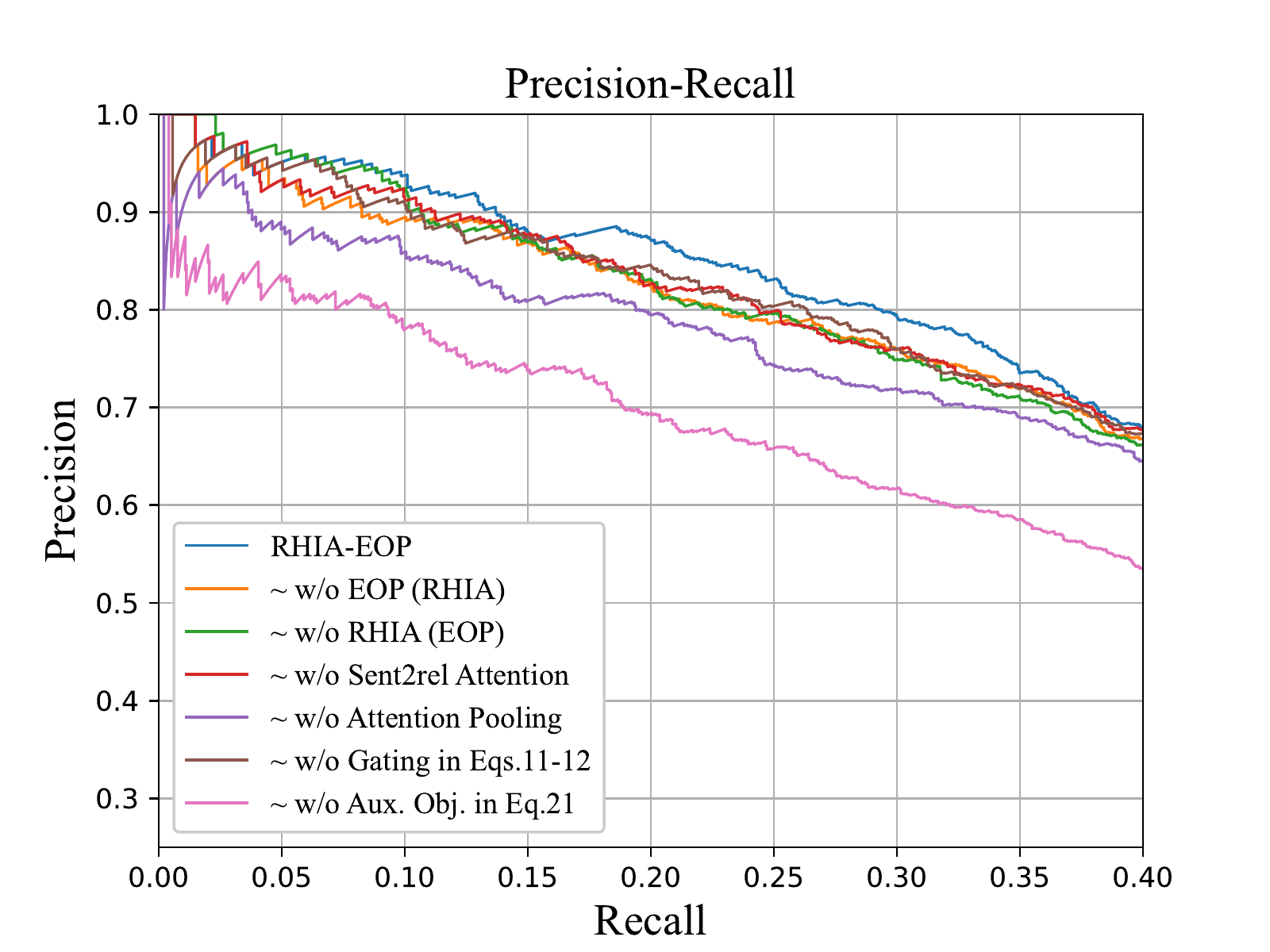}
    \end{minipage}
  }
  \caption{Precision-recall (PR) curves.}
  \label{fig:pr}
\end{figure*}

The comparison results of different models are displayed in Table~\ref{tab:table2}, Table~\ref{tab:table3} and Figure~\ref{fig:pr} (a). Note that, the results in Table~\ref{tab:table2} are obtained using the full test set of the NYT dataset. While the results in Table ~\ref{tab:table3} are obtained using the remaining test set with all single-sentence bags removed as \citep{li2020improving}, because for each single-sentence bag, the result is the same whether one, two or all sentences are retained for evaluation. Here the single-sentence bag means a bag consisting of only one sentence.

\begin{table*}[!ht]
	\centering
  \caption{Model evaluation and ablation study on NYT. The best scores are \textbf{bolded} or \textbf{underlined} in model comparison and ablation respectively. }
	\label{tab:table2}
	\resizebox{\linewidth}{!}{
	  \begin{tabular}{lccccccccc}
		  \toprule

		  Approach  & P@100 & P@200 & P@300 & P@500 & P@1000 & P@2000 & Mean & AUC & Max\_F1 \\
          \midrule
		  PCNN\_ATT \citep{lin2016neural}  & 78.0  & 72.5  & 71.0  & 67.6  & 54.3   & 40.8   & 64.0 & 0.39 & 0.437 \\
		  HNRE \citep{han2018hierarchical}   & 82.0  & 80.5  & 76.0  & 67.8  & 58.3   & 42.1   & 67.8 & 0.42& 0.455 \\
		  ToHRE \citep{yu2020tohre}   & 91.5  & 82.9  & 79.6  & 74.8  & 63.3   & 48.9   & 73.5 & 0.44 & 0.476 \\
		  CoRA \citep{li2020improving}   & 93.0  & 91.0 & 88.0  & 81.2  & 67.6   & 51.4   & 78.7 & 0.53 & 0.525 \\
      HNRE w/ Ent. Emb. & 84.7 & 83.1 & 77.3 & 75.5 & 65.6 & 47.9 & 72.4 & 0.47 & 0.467 \\
      HNRE w/ Aux. Obj. & 85.0 & 82.5 & 79.3 & 73.9 & 64.2 & 49.8 & 72.5 & 0.49 & 0.485 \\
      HNRE w/ Ent. Emb. w/ Aux. Obj. & 90.1 & 87.0 & 85.7 & 76.8 & 67.3 & 51.7 & 76.4 & 0.52 & 0.525 \\
		  \midrule
		  \textbf{RHIA-EOP}   & \textbf{95.0}  & \textbf{94.0}  & \textbf{89.7} & \textbf{85.2}  & \textbf{71.7}  & \textbf{53.2}   & \textbf{81.5}  & \textbf{0.56} & \textbf{0.546}  \\
          \midrule
          \emph{Ablations} & & & & & & & & & \\
          \midrule
          $\sim$ w/o EOP (RHIA)   & 93.0  & 89.0  & 88.3 & 81.0  & 70.8  & 52.1  & 79.0 & 0.546 & 0.531  \\
          $\sim$ w/o RHIA (EOP)  & \underline{96.0}  & 93.0  & 88.3 & 81.2  & 70.3  & 51.8  & 80.1 & 0.545 & 0.529  \\
          $\sim$ w/o Sent2rel Attention & 93.0 & 92.5 & 89.0 & 82.2 & 71.2 & 53.0 & 80.2 & 0.548 & 0.539 \\
          $\sim$ w/o Attention Pooling & 89.0 & 87.0 & 83.3 & 79.6 & 68.9 & 52.0 & 76.6 & 0.534 & 0.531 \\
          $\sim$ w/o Gating in Eqs.\ref{Eq-11}-\ref{Eq-12} & 95.0 & 91.0 & 87.3 & 83.2 & 70.8 & 53.0 & 80.1 & 0.549 & 0.541 \\
          $\sim$ w/o Aux. Obj. in Eq.\ref{Eq-21} & 81.5 & 81.0 & 76.0 & 71.6 & 60.8 & 47.2 & 69.7 & 0.451 & 0.481 \\
          $\sim$ w/ uncased BERT-Base & 91.5 & 91.0 & 88.1 & 84.4 & 69.9 & 52.5 & 79.6 & \underline{0.593} & \underline{0.584} \\
		  \bottomrule
	  \end{tabular}
    }
    
\end{table*}

\begin{table*}[!ht]
	\centering
  \caption{Model evaluation and ablation study on NYT when retaining one/two/all sentence(s) in each bag at random. The best scores are \textbf{bolded} or \textbf{underlined} in model comparison and ablation respectively.}
	\label{tab:table3}
	\resizebox{\linewidth}{!}{
	  \begin{tabular}{lccccccccccccccc}
		  \toprule
		  \ & \multicolumn{4}{c}{One} & \multicolumn{4}{c}{Two} & \multicolumn{4}{c}{All} \\
		  
		  \textbf{P@N(\%)}  & 100 & 200 & 300 & Mean & 100 & 200 & 300 & Mean & 100 & 200 & 300 & Mean \\
          \midrule
		  PCNN\_ATT \citep{lin2016neural} & 73.3 & 69.2 & 60.8 & 67.8 & 77.2 & 71.6 & 66.1 & 71.6 & 76.2 & 73.1 & 67.4 & 72.2 \\
		  HNRE \citep{han2018hierarchical} & 84.0 & 76.0 & 69.7 & 76.6 & 85.0 & 76.0 & 72.7 & 77.9 & 88.0 & 79.5 & 75.3 & 80.9 \\
	    ToHRE \citep{yu2020tohre} & 87.1 & 81.4 & 75.3 & 81.3 & 89.7 & 83.1 & 78.5 & 83.8 & 92.4 & 86.7 & 81.2 & 86.8 \\
		  CoRA \citep{li2020improving} & 94.0 & 90.5 & 82.0 & 88.8 &  \textbf{98.0} & 91.0 & 86.3 & 91.8 & \textbf{98.0} & 92.5 & 88.3 & 92.9 \\
      HNRE w/ Ent. Emb. & 90.2 & 86.5 & 81.3 & 86.0 & 91.3 & 87.2 & 82.4 & 87.0 & 93.1 & 89.0 & 85.8 & 89.3 \\
      HNRE w/ Aux. Obj. & 87.7 & 82.0 & 76.8 & 82.2 & 88.0 & 84.4 & 79.2 & 83.9 & 91.0 & 85.5 & 82.7 & 86.4 \\
      HNRE w/ Ent. Emb. w/ Aux. Obj. & 93.0 & 85.4 & 81.9 & 86.8 & 94.0 & 90.1 & 84.7 & 89.6 & 94.0 & 90.9 & 87.5 & 90.8 \\ 
		  \midrule
		  \textbf{RHIA-EOP} & \textbf{96.0} & \textbf{92.5} & \textbf{86.7} & \textbf{91.7} & \textbf{98.0} & \textbf{95.5} & \textbf{92.3} & \textbf{95.3} & \textbf{98.0} & \textbf{96.5} & \textbf{93.3} & \textbf{95.9} \\
      
          \bottomrule
         \emph{Ablations} & & & & & & & & & & \\
          \midrule
          $\sim$ w/o EOP (RHIA) & 95.0 & 89.5 & 84.0 & 89.5 & 96.0 & 95.0 & 90.3 & 93.8 & 97.0 & \underline{96.5} & 91.0 & 94.8 \\
          $\sim$ w/o RHIA (EOP) & 95.0 & 91.0 & \underline{86.7} & 90.9 & \underline{99.0} & 92.0 & 87.7 & 92.9 & \underline{99.0} & 94.5 & 90.0 & 94.5 \\
          $\sim$ w/o Sent2rel Attention & 93.3 & 91.0 & 85.0 & 89.8 & 95.0 & 92.0 & 89.3 & 92.1 & 96.0 & 94.0 & 91.7 & 93.9 \\
          $\sim$ w/o Attention Pooling & 92.0 & 90.0 & 87.7 & 89.9 & 93.0 & 93.0 & 90.7 & 92.2 & 94.0 & 94.0 & 91.3 & 93.1 \\
          $\sim$ w/o Gating in Eqs.\ref{Eq-11}-\ref{Eq-12} & 95.0 & 92.5 & 86.0 & 91.2 & 96.0 & 94.5 & 90.7 & 93.7 & 96.0 & 95.0 & 93.3 & 94.8 \\
          $\sim$ w/o Aux. Obj. in Eq.\ref{Eq-21} & 83.0 & 81.0 & 74.0 & 79.3 & 89.0 & 83.5 & 78.3 & 83.6 & 89.0 & 85.5 & 80.7 & 85.1 \\
          $\sim$ w/ uncased BERT-Base & 94.0 & 90.4 & 85.3 & 89.9 & 95.2 & 93.5 & 90.0 & 92.9 & 96.4 & 93.8 & 91.9 & 94.0\\
		  \bottomrule
	  \end{tabular}
    }
    
\end{table*} 

It can be observed that RHIA-EOP significantly 
outperforms many baseline models in all metrics at the same time.
Our approach achieves the AUC of 0.56, which outperforms the strong baseline CoRA (0.53) by 0.03. For Max\_F1, we improve baseline approaches by at least 2.1\%. For Top-N precision metric, we take relatively more unique $N$ values in Table~\ref{tab:table2}, and find that RHIA-EOP sets the best scores on all $N$ values. And in another setting, i.e., randomly retaining one, two or all sentence(s) in each bag, RHIA-EOP achieves the best results despite the randomness of retained sentences in Table~\ref{tab:table3} (Almost all values exceed 90\%). Besides, the curve of RHIA-EOP is significantly higher than PCNN\_ATT, HNRE, ToHRE and CoRA. The results of setups \textbf{HNRE w/ Ent. Emb.} and \textbf{HNRE w/ Aux. Obj.} indicate that both the Entity-Aware Embedding layer and the hierarchical auxiliary objective bring substantial performance improvements. However, the results of \textbf{HNRE w/ Ent. Emb. w/ Aux. Obj.} are still worse than our RHIA-EOP because the heuristic effects between relation levels and the entity order information are still not considered. 

To measure RHIA-EOP's ability to handle long-tail relations, we conduct a model comparison on all long-tail relations, and the results are shown in table~\ref{tab:table4}.
Here Hits@K is employed to measure this ability, which indicates whether the ground-label's probability of a bag ranking in the top-K relations. During the calculation, we use macro average regarding different relations.
Very inspiringly, RHIA-EOP achieve the highest results at all values of $K$. These results confirm the heuristic influence of higher-level relations on the lower-level ones and the advantage of relation hierarchies.

\begin{table}[!htbp]
  \caption{Hits@K on all long-tail relations. Here the relation is long-tail if its number of instances is less than 100 or 200.}
    \label{tab:table4}
  \centering
  \resizebox{0.8\linewidth}{!}{
    \begin{tabular}{lccccccc}
        \toprule
        \#Instance & \ & \textless 100 & \ & \ & \textless 200 & \ & \\
        Hits@K & 10 & 15 & 20 & 10 & 15 & 20 & \\
        \midrule
        PCNN\_ATT \citep{lin2016neural} & \textless 5.0 & 7.4 & 40.7 & 17.2 & 24.2 & 51.5 \\
        HNRE \citep{han2018hierarchical} & 29.6 & 51.9 & 61.1 & 41.4 & 60.6 & 68.2 \\
        ToHRE \citep{yu2020tohre} & 62.9 & 75.9 & 81.4 & 69.7 & 80.3 & 84.8 \\
        CoRA \citep{li2020improving} & \textbf{66.7} & 72.2 & 87.0 & \textbf{72.7} & 77.3 & 89.3 \\
        HNRE w/ Ent. Emb. & 36.4 & 52.2 & 69.6 & 54.5 & 63.6 & 77.8 \\
        HNRE w/ Aux. Obj. & 44.4 & 54.5 & 68.2 & 60.6 & 72.4 & 81.8 \\
        HNRE w/ Ent. Emb. w/ Aux. Obj. & 47.8 & 62.9 & 77.3 & 66.7 & 75.9 & 87.0 \\
        \midrule
        RHIA-EOP & \textbf{66.7} & \textbf{83.3} & \textbf{94.4} & \textbf{72.7} & \textbf{86.4} & \textbf{95.5} \\
        
        $\sim$ w/ uncased BERT-Base & 55.6 & 66.7 & 72.2 & 63.6 & 77.3 & 81.8 \\
       
        \bottomrule
    \end{tabular}
  }
\end{table}

In addition, we briefly analyze the computational overhead. Both our RHIA-EOP and the CORA of \citet{li2020improving} employ multiple attention mechanisms. We further introduce the interaction effects between relation levels and entity order information, and the number of parameters grow from 13M to 18M. It is still much smaller compared to the pre-trained language models since the smallest BERT-Base has 110M parameters). The parameter growth is not really huge (i.e., 5M). The required training time dose not increase dramatically, both models can converge within two hours, while the performance gains are relatively huge and substantial. Therefore, although some computational overhead is added, it is worthwhile.

\subsection{Ablation Study}

In order to validate the effectiveness of each module in RHIA-EOP and to analyze the sources of performance improvement, we evaluate the following ablation experimental setups:

\begin{itemize}
	\item \textbf{$\sim$ w/o EOP (RHIA)}: The Entity-Order Perception subtask is removed. It is equivalent to having only the Recursive Hierarchy-Interactive Attention (RHIA) module.
	\item \textbf{$\sim$ w/o RHIA (EOP)}: The module RHIA is replaced by CoRA \citep{li2020improving}. It is equivalent to combining CoRA and EOP.
	\item \textbf{$\sim$ w/o Sent2rel Attention}: The sentence-to-relation attention (i.e., Eqs.\ref{Eq-5}-\ref{Eq-10}) is replaced by average pooling.
	\item \textbf{$\sim$ w/o Attention Pooling}: The attention pooling (i.e., Eq.\ref{Eq-17}) is replaced by average pooling.
	\item \textbf{$\sim$ w/o Gating in Eqs.\ref{Eq-11}-\ref{Eq-12}}: Eqs.\ref{Eq-11}-\ref{Eq-12} are removed. Feature concatenation $[u; \hat{c}^{(i)}]$ is directly passed into MultiLayer Perceptron (MLP).
	\item \textbf{$\sim$ w/o Aux. Obj. in Eq.\ref{Eq-21}}: The hierarchical auxiliary objective $L_{hier}$ is removed (i.e., Eq. \ref{Eq-21}).
	\item \textbf{$\sim$ w/ uncased BERT-Base}: The Entity-Aware Embedding layer is replaced by uncased BERT-Base model. 
\end{itemize}

The results are shown in the bottom of Table~\ref{tab:table2} and Table\ref{tab:table3}, and Figure~\ref{fig:pr} (b).
And the evaluation results reflect consistent declines in P@N, Max\_F1 and AUC. For the two main modules of RHIA-EOP, in Table~\ref{tab:table2}, compared to RHIA-EOP, both RHIA and EOP drop almost 0.015 on AUC, and these two models decrease by 2.5\% and 1.4\% on the mean of P@N, respectively. While in Table~\ref{tab:table3}, the performance drop is similar obviously, the mean of P@N of RHIA decreases by 2.2\%, 1.5\% and 1.1\%, respectively. For EOP, the values are 0.8\%, 2.4\% and 1.4\%. 

To further analyze whether the performance improvement comes from the new attention mechanism or from the full utilization of relation hierarchies, we set up more experimental settings. The setups \textbf{$\sim$ w/o Sent2rel Attention} and \textbf{$\sim$ w/o Attention Pooling} argue for the importance and validity of the attention mechanism.
As for the relation hierarchies, 
firstly, for each level of the relation hierarchies, we construct the corresponding feature representation for each bag and classify it. This is the hierarchical auxiliary objective (i.e., Eq.\ref{Eq-21}). The setup \textbf{$\sim$ w/o Aux. Obj. in Eq.\ref{Eq-21}} has demonstrated that not using the relation hierarchies leads to a significant performance drop.
Secondly, along the relation hierarchies, relational information is propagated in a recursive form (i.e., RHIA). The setup \textbf{$\sim$ w/o EOP (RHIA)} confirms its effectiveness.
To sum up, the relation hierarchies are indeed contributing and the attention mechanism is indeed effective.

Interestingly, despite the Pre-trained Language Models (PLMs) are so powerful, the results achieved by the BERT-based implementation are not satisfactory. Except for the values of AUC and Max\_F1, the values of all metrics have decreased. The reason may be that BERT cannot highlight two entities as the Entity-Aware Embedding layer does, i.e., it ignores the importance of the entity-pair itself and the position of entities. After all, the essence of DSRE task lies in the classification of entity pairs.

\subsection{Statistical Analysis of Multiple training runs}
\label{subsec:sta}

\begin{table}[!tbp]
  \centering
  \caption{The mean and standard deviation of RHIA-EOP.}
  \label{tab:sta2}
  \resizebox{0.8\linewidth}{!}{
    \begin{tabular}{lcccccccc}
        \toprule
        Metrics & P@N (Mean) & AUC & Max\_F1\\
        \midrule
        Values & 80.82 $\pm$ 0.676 & 0.561 $\pm$ 0.003 & 0.546 $\pm$ 0.009 \\
        \bottomrule  
    \end{tabular}
  }
\end{table}

\begin{table}[!tbp]
  \centering
  \caption{The mean and standard deviation of RHIA-EOP on NYT when retaining one/two/all sentence(s) in each bag at random.}
  \label{tab:sta3}
  \resizebox{0.8\linewidth}{!}{
    \begin{tabular}{lcccccccc}
        \toprule
        P@N (Mean) & ONE & TWO & ALL\\
        \midrule
        Values & 91.86 $\pm$ 0.873 & 94.36 $\pm$ 0.653 & 95.66 $\pm$ 0.224 \\ 
        \bottomrule  
    \end{tabular}
  }
\end{table}

We reran our model five times to calculate the respective means and standard deviations in terms of AUC, Max\_F1 and the mean of P@N. Here for P@N, the randomness of the selected samples when retaining one/two/all sentence(s) in each bag at random leads to large fluctuations in its value, so we only report the mean and standard deviation of ``the mean of P@N", i.e., P@N (Mean).
Detailed results can be found in Table~\ref{tab:sta2} and Table~\ref{tab:sta3}. These results demonstrate that the performance gains are stable and convincing.

\subsection{Case Study}

To visualize the heuristic effect between relation hierarchies, two example sentences from NYT are selected. And then  we list/analyze the Top-3 attention scores at all relation levels of these two sentences in Table~\ref{tab:table8}. Our RHIA-EOP, compared to CoRA, better handles the highly unbalanced category (i.e., \textbf{NA}), and gives greater scores to \textbf{NA} for the true instances of \textbf{NA} and less scores to \textbf{NA} for other instances. This capability can also improve the recognition of noisy sentences from the side, and alleviate the wrong labeling problem. These findings demonstrate the effectiveness of RHIA-EOP.

\begin{table*}[!htbp]
  \centering
  \caption{The Top-3 attention scores in Eq.\ref{Eq-5} at all relation levels of two examples from NYT. For the selected relation fact, Example 1 is noisy, while Example 2 is correctly labeled.}
  \label{tab:table8}
  \resizebox{\textwidth}{!}{
    \begin{tabular}{lllllll}
        \toprule
        Relation fact: & \multicolumn{6}{l}{\textless \textbf{Grameen Bank}, \emph{/business/company/founders}, \textbf{Muhammad Yunus} \textgreater} \\
        \bottomrule	
        Example 1: & \multicolumn{6}{l}{\textbf{Muhammad Yunus}, who won the Nobel Peace Prize, last year, demonstrated with,} \\
        \ & \multicolumn{6}{l}{\textbf{Grameen Bank} the power of microfinancing.} \\
        Example 2: & \multicolumn{6}{l}{On Sunday, though, there was a significant shift of the tectonic plates of Bangladeshi politics,} \\
        \ & \multicolumn{6}{l}{as \textbf{Muhammad Yunus}, the founder of a microfinance empire, known as the \textbf{Grameen Bank} and} \\
        \ & \multicolumn{6}{l}{the winner of the 2006 Nobel Peace Prize, announced that he would start a new party and step} \\
        \ & \multicolumn{6}{l}{into the electoral fray.} \\
        \midrule
        \midrule
        Example 1 & \multicolumn{2}{c}{$\alpha^{(1)}$} & \multicolumn{2}{c}{$\alpha^{(2)}$} & \multicolumn{2}{c}{$\alpha^{(3)}$} \\
        \midrule
        \ & \emph{/business}: & 0.422 & \emph{NA}: & 0.383 & \emph{NA}: & 0.387 \\
        CoRA & \emph{NA}: & 0.384 & \emph{/business/company}: & 0.272 & \emph{/business/company/founders}: & 0.197 \\
        \ & \emph{/location}: & 0.037 & \emph{/business/person}: & 0.063 & \emph{/business/person/company}: & 0.063 \\
        \hline
        \ & \emph{NA}: & 0.524 & \emph{NA}: & 0.612 & \emph{NA}: & 0.559 \\
        RHIA-EOP & \emph{/business}: & 0.324 & \emph{/business/company}: & 0.095 & \emph{/business/person/company}: & 0.089 \\
        \ & \emph{/people}: & 0.049 & \emph{/business/person}: & 0.072 & \emph{/business/company/founders}: & 0.082 \\
        \bottomrule
        \toprule
        Example 2 & \multicolumn{2}{c}{$\alpha^{(1)}$} & \multicolumn{2}{c}{$\alpha^{(2)}$} & \multicolumn{2}{c}{$\alpha^{(3)}$} \\
        \midrule
        & \emph{/business}: & 0.755 & \emph{/business/company}: & 0.679 & \emph{/business/company/founders}: & 0.652 \\
        CoRA & \emph{NA}: & 0.103 & \emph{NA}: & 0.089 & \emph{NA}: & 0.069 \\
        \ & \emph{/people}: & 0.031 & \emph{/business/person}: & 0.059 & \emph{/business/person/company}: & 0.057 \\
        \midrule
        \ & \emph{/business}: & 0.899 & \emph{/business/company}: & 0.856 & \emph{/business/company/founders}: & 0.741 \\
        RHIA-EOP & \emph{NA}: & 0.024 & \emph{/business/person}: & 0.047 & \emph{/business/person/company}: & 0.052 \\
        \ & \emph{/people}: & 0.022 & \emph{NA}: & 0.019 & \emph{/business/company/major\_shareholders}: & 0.044 \\
        \bottomrule
    \end{tabular}
  }
\end{table*}

\section{Conclusions}

In this paper, we fully exploit the inherent taxonomic structure of relations and design a recursive hierarchy-interactive attention network. In this way, we model the heuristic influence of higher-level relations on the lower-level ones. Furthermore, multiple training objectives are designed to take entity-order information into account. The substantial experiments on the NYT dataset show that RHIA-EOP achieves state-of-the-art performance in multiple metrics, including standard metrics (i.e., AUC, P-R curve, Top-N precision, etc.) and long-tail metrics (i.e., Hits@K). 
In the future, we plan to consider the interaction between the many sibling relations within the same level and employ our recursive approach to other tasks, such as fine-grained hierarchical text classification. Besides, we will further explore the extension of traditional feature-based methods in the deep learning paradigm.

\section*{CRediT authorship contribution statement}

Ridong Han: Methodology, Conceptualization, Software, Visualization, Writing, Editing. 
Tao Peng: Supervision, Funding acquisition, Reviewing, Validation.
Jiayu Han: Editing, Reviewing. 
Hai Cui: Formal analysis, Validation. 
Lu Liu: Reviewing, Supervision, Validation, Funding
acquisition.

\section*{Acknowledgment}
This work is supported by the National Natural Science Foundation of China under grant No.61872163 and 61806084, Jilin Province Key Scientific and Technological Research and Development Project under grant No.20210201131GX, and Jilin Provincial Education Department Project under grant No.JJKH20190160KJ.

\bibliography{mybibfile}

\begin{thebibliography}{44}
\expandafter\ifx\csname natexlab\endcsname\relax\def\natexlab#1{#1}\fi
\providecommand{\url}[1]{\texttt{#1}}
\providecommand{\href}[2]{#2}
\providecommand{\path}[1]{#1}
\providecommand{\DOIprefix}{doi:}
\providecommand{\ArXivprefix}{arXiv:}
\providecommand{\URLprefix}{URL: }
\providecommand{\Pubmedprefix}{pmid:}
\providecommand{\doi}[1]{\href{http://dx.doi.org/#1}{\path{#1}}}
\providecommand{\Pubmed}[1]{\href{pmid:#1}{\path{#1}}}
\providecommand{\bibinfo}[2]{#2}
\ifx\xfnm\relax \def\xfnm[#1]{\unskip,\space#1}\fi
\bibitem[{Suchanek et~al.(2007)Suchanek, Kasneci, and Weikum}]{Fabian2007yago}
\bibinfo{author}{F.~M. Suchanek}, \bibinfo{author}{G.~Kasneci},
  \bibinfo{author}{G.~Weikum},
\newblock \bibinfo{title}{Yago: a core of semantic knowledge},
\newblock in: \bibinfo{booktitle}{Proceedings of the 16th International
  Conference on World Wide Web, {WWW} 2007}, \bibinfo{publisher}{{ACM}},
  \bibinfo{year}{2007}, pp. \bibinfo{pages}{697--706}.
  \DOIprefix\doi{10.1145/1242572.1242667}.
\bibitem[{Bollacker et~al.(2008)Bollacker, Evans, Paritosh, Sturge, and
  Taylor}]{kurt2008freebase}
\bibinfo{author}{K.~D. Bollacker}, \bibinfo{author}{C.~Evans},
  \bibinfo{author}{P.~Paritosh}, \bibinfo{author}{T.~Sturge},
  \bibinfo{author}{J.~Taylor},
\newblock \bibinfo{title}{Freebase: a collaboratively created graph database
  for structuring human knowledge},
\newblock in: \bibinfo{booktitle}{Proceedings of the {ACM} {SIGMOD}
  International Conference on Management of Data, {SIGMOD} 2008},
  \bibinfo{publisher}{{ACM}}, \bibinfo{year}{2008}, pp.
  \bibinfo{pages}{1247--1250}. \DOIprefix\doi{10.1145/1376616.1376746}.
\bibitem[{Lehmann et~al.(2015)Lehmann, Isele, Jakob, Jentzsch, Kontokostas,
  Mendes, Hellmann, Morsey, van Kleef, Auer, and Bizer}]{jens2015dbpedia}
\bibinfo{author}{J.~Lehmann}, \bibinfo{author}{R.~Isele},
  \bibinfo{author}{M.~Jakob}, \bibinfo{author}{A.~Jentzsch},
  \bibinfo{author}{D.~Kontokostas}, \bibinfo{author}{P.~N. Mendes},
  \bibinfo{author}{S.~Hellmann}, \bibinfo{author}{M.~Morsey},
  \bibinfo{author}{P.~van Kleef}, \bibinfo{author}{S.~Auer},
  \bibinfo{author}{C.~Bizer},
\newblock \bibinfo{title}{Dbpedia - {A} large-scale, multilingual knowledge
  base extracted from wikipedia},
\newblock \bibinfo{journal}{Semantic Web} \bibinfo{volume}{6}
  (\bibinfo{year}{2015}) \bibinfo{pages}{167--195}.
  \DOIprefix\doi{10.3233/SW-140134}.
\bibitem[{Mintz et~al.(2009)Mintz, Bills, Snow, and Jurafsky}]{mike2009distant}
\bibinfo{author}{M.~Mintz}, \bibinfo{author}{S.~Bills},
  \bibinfo{author}{R.~Snow}, \bibinfo{author}{D.~Jurafsky},
\newblock \bibinfo{title}{Distant supervision for relation extraction without
  labeled data},
\newblock in: \bibinfo{booktitle}{Proceedings of the 47th Annual Meeting of the
  Association for Computational Linguistics and the 4th International Joint
  Conference on Natural Language Processing of the AFNLP, {ACL} 2009},
  \bibinfo{publisher}{The Association for Computer Linguistics},
  \bibinfo{year}{2009}, pp. \bibinfo{pages}{1003--1011}.
\bibitem[{Riedel et~al.(2010)Riedel, Yao, and McCallum}]{riedel2010modeling}
\bibinfo{author}{S.~Riedel}, \bibinfo{author}{L.~Yao},
  \bibinfo{author}{A.~McCallum},
\newblock \bibinfo{title}{Modeling relations and their mentions without labeled
  text},
\newblock in: \bibinfo{booktitle}{Machine Learning and Knowledge Discovery in
  Databases, European Conference, {ECML} {PKDD} 2010}, volume
  \bibinfo{volume}{6323} of \textit{\bibinfo{series}{Lecture Notes in Computer
  Science}}, \bibinfo{publisher}{Springer}, \bibinfo{year}{2010}, pp.
  \bibinfo{pages}{148--163}. \DOIprefix\doi{10.1007/978-3-642-15939-8\_10}.
\bibitem[{Hoffmann et~al.(2011)Hoffmann, Zhang, Ling, Zettlemoyer, and
  Weld}]{hoffmann2011knowledge}
\bibinfo{author}{R.~Hoffmann}, \bibinfo{author}{C.~Zhang},
  \bibinfo{author}{X.~Ling}, \bibinfo{author}{L.~S. Zettlemoyer},
  \bibinfo{author}{D.~S. Weld},
\newblock \bibinfo{title}{Knowledge-based weak supervision for information
  extraction of overlapping relations},
\newblock in: \bibinfo{booktitle}{Proceedings of the 49th annual meeting of the
  association for computational linguistics: human language technologies},
  \bibinfo{publisher}{The Association for Computer Linguistics},
  \bibinfo{year}{2011}, pp. \bibinfo{pages}{541--550}.
\bibitem[{Lin et~al.(2016)Lin, Shen, Liu, Luan, and Sun}]{lin2016neural}
\bibinfo{author}{Y.~Lin}, \bibinfo{author}{S.~Shen}, \bibinfo{author}{Z.~Liu},
  \bibinfo{author}{H.~Luan}, \bibinfo{author}{M.~Sun},
\newblock \bibinfo{title}{Neural relation extraction with selective attention
  over instances},
\newblock in: \bibinfo{booktitle}{Proceedings of the 54th Annual Meeting of the
  Association for Computational Linguistics, {ACL} 2016},
  \bibinfo{publisher}{Association for Computational Linguistics},
  \bibinfo{address}{Berlin, Germany}, \bibinfo{year}{2016}, pp.
  \bibinfo{pages}{2124--2133}. \DOIprefix\doi{10.18653/v1/P16-1200}.
\bibitem[{Qu et~al.(2018)Qu, Ouyang, Hua, Ye, and Li}]{qu2018distant}
\bibinfo{author}{J.~Qu}, \bibinfo{author}{D.~Ouyang}, \bibinfo{author}{W.~Hua},
  \bibinfo{author}{Y.~Ye}, \bibinfo{author}{X.~Li},
\newblock \bibinfo{title}{Distant supervision for neural relation extraction
  integrated with word attention and property features},
\newblock \bibinfo{journal}{Neural Networks} \bibinfo{volume}{100}
  (\bibinfo{year}{2018}) \bibinfo{pages}{59--69}.
  \DOIprefix\doi{10.1016/j.neunet.2018.01.006}.
\bibitem[{Du et~al.(2018)Du, Han, Way, and Wan}]{du2018multi}
\bibinfo{author}{J.~Du}, \bibinfo{author}{J.~Han}, \bibinfo{author}{A.~Way},
  \bibinfo{author}{D.~Wan},
\newblock \bibinfo{title}{Multi-level structured self-attentions for distantly
  supervised relation extraction},
\newblock in: \bibinfo{booktitle}{Proceedings of the 2018 Conference on
  Empirical Methods in Natural Language Processing, {EMNLP} 2018},
  \bibinfo{publisher}{Association for Computational Linguistics},
  \bibinfo{year}{2018}, pp. \bibinfo{pages}{2216--2225}.
  \DOIprefix\doi{10.18653/v1/d18-1245}.
\bibitem[{Liu et~al.(2017)Liu, Wang, Chang, and Sui}]{liu2017softlabel}
\bibinfo{author}{T.~Liu}, \bibinfo{author}{K.~Wang},
  \bibinfo{author}{B.~Chang}, \bibinfo{author}{Z.~Sui},
\newblock \bibinfo{title}{A soft-label method for noise-tolerant distantly
  supervised relation extraction},
\newblock in: \bibinfo{booktitle}{Proceedings of the 2017 Conference on
  Empirical Methods in Natural Language Processing, {EMNLP} 2017},
  \bibinfo{publisher}{Association for Computational Linguistics},
  \bibinfo{year}{2017}, pp. \bibinfo{pages}{1790--1795}.
  \DOIprefix\doi{10.18653/v1/d17-1189}.
\bibitem[{Xiao et~al.(2020)Xiao, Tan, Fan, Xu, and Zhu}]{xiao2020joint}
\bibinfo{author}{Y.~Xiao}, \bibinfo{author}{C.~Tan}, \bibinfo{author}{Z.~Fan},
  \bibinfo{author}{Q.~Xu}, \bibinfo{author}{W.~Zhu},
\newblock \bibinfo{title}{Joint entity and relation extraction with a hybrid
  transformer and reinforcement learning based model},
\newblock in: \bibinfo{booktitle}{Proceedings of the Thirty-Fourth {AAAI}
  Conference on Artificial Intelligence, {AAAI} 2020},
  \bibinfo{publisher}{{AAAI} Press}, \bibinfo{year}{2020}, pp.
  \bibinfo{pages}{9314--9321}.
\bibitem[{Yang et~al.(2020)Yang, Wang, Su, and Wang}]{yang2020threat}
\bibinfo{author}{J.~Yang}, \bibinfo{author}{Q.~Wang}, \bibinfo{author}{C.~Su},
  \bibinfo{author}{X.~Wang},
\newblock \bibinfo{title}{Threat intelligence relationship extraction based on
  distant supervision and reinforcement learning {(S)}},
\newblock in: \bibinfo{booktitle}{Proceedings of the 32nd International
  Conference on Software Engineering and Knowledge Engineering, {SEKE} 2020},
  \bibinfo{publisher}{{KSI} Research Inc.}, \bibinfo{year}{2020}, pp.
  \bibinfo{pages}{572--576}. \DOIprefix\doi{10.18293/SEKE2020-149}.
\bibitem[{Han et~al.(2018)Han, Yu, Liu, Sun, and Li}]{han2018hierarchical}
\bibinfo{author}{X.~Han}, \bibinfo{author}{P.~Yu}, \bibinfo{author}{Z.~Liu},
  \bibinfo{author}{M.~Sun}, \bibinfo{author}{P.~Li},
\newblock \bibinfo{title}{Hierarchical relation extraction with coarse-to-fine
  grained attention},
\newblock in: \bibinfo{booktitle}{Proceedings of the 2018 Conference on
  Empirical Methods in Natural Language Processing, {EMNLP} 2018},
  \bibinfo{publisher}{Association for Computational Linguistics},
  \bibinfo{year}{2018}, pp. \bibinfo{pages}{2236--2245}.
  \DOIprefix\doi{10.18653/v1/d18-1247}.
\bibitem[{Zhang et~al.(2019)Zhang, Deng, Sun, Wang, Chen, Zhang, and
  Chen}]{zhang2019long}
\bibinfo{author}{N.~Zhang}, \bibinfo{author}{S.~Deng},
  \bibinfo{author}{Z.~Sun}, \bibinfo{author}{G.~Wang},
  \bibinfo{author}{X.~Chen}, \bibinfo{author}{W.~Zhang},
  \bibinfo{author}{H.~Chen},
\newblock \bibinfo{title}{Long-tail relation extraction via knowledge graph
  embeddings and graph convolution networks},
\newblock in: \bibinfo{booktitle}{Proceedings of the 2019 Conference of the
  North American Chapter of the Association for Computational Linguistics:
  Human Language Technologies, {NAACL-HLT} 2019},
  \bibinfo{publisher}{Association for Computational Linguistics},
  \bibinfo{year}{2019}, pp. \bibinfo{pages}{3016--3025}.
  \DOIprefix\doi{10.18653/v1/n19-1306}.
\bibitem[{Li et~al.(2020)Li, Shen, Long, Jiang, Zhou, and
  Zhang}]{li2020improving}
\bibinfo{author}{Y.~Li}, \bibinfo{author}{T.~Shen}, \bibinfo{author}{G.~Long},
  \bibinfo{author}{J.~Jiang}, \bibinfo{author}{T.~Zhou},
  \bibinfo{author}{C.~Zhang},
\newblock \bibinfo{title}{Improving long-tail relation extraction with
  collaborating relation-augmented attention},
\newblock in: \bibinfo{booktitle}{Proceedings of the 28th International
  Conference on Computational Linguistics, {COLING} 2020},
  \bibinfo{publisher}{International Committee on Computational Linguistics},
  \bibinfo{year}{2020}, pp. \bibinfo{pages}{1653--1664}.
  \DOIprefix\doi{10.18653/v1/2020.coling-main.145}.
\bibitem[{Yu et~al.(2020)Yu, Han, Tian, and Chang}]{yu2020tohre}
\bibinfo{author}{E.~Yu}, \bibinfo{author}{W.~Han}, \bibinfo{author}{Y.~Tian},
  \bibinfo{author}{Y.~Chang},
\newblock \bibinfo{title}{Tohre: {A} top-down classification strategy with
  hierarchical bag representation for distantly supervised relation
  extraction},
\newblock in: \bibinfo{booktitle}{Proceedings of the 28th International
  Conference on Computational Linguistics, {COLING} 2020},
  \bibinfo{publisher}{International Committee on Computational Linguistics},
  \bibinfo{year}{2020}, pp. \bibinfo{pages}{1665--1676}.
  \DOIprefix\doi{10.18653/v1/2020.coling-main.146}.
\bibitem[{Bordes et~al.(2013)Bordes, Usunier, Garc{\'{\i}}a{-}Dur{\'{a}}n,
  Weston, and Yakhnenko}]{bordes2013transe}
\bibinfo{author}{A.~Bordes}, \bibinfo{author}{N.~Usunier},
  \bibinfo{author}{A.~Garc{\'{\i}}a{-}Dur{\'{a}}n},
  \bibinfo{author}{J.~Weston}, \bibinfo{author}{O.~Yakhnenko},
\newblock \bibinfo{title}{Translating embeddings for modeling multi-relational
  data},
\newblock in: \bibinfo{booktitle}{Proceedings of the 27th Annual Conference on
  Neural Information Processing Systems, {NIPS} 2013}, \bibinfo{year}{2013},
  pp. \bibinfo{pages}{2787--2795}.
\bibitem[{Defferrard et~al.(2016)Defferrard, Bresson, and
  Vandergheynst}]{defferrard2016convolutional}
\bibinfo{author}{M.~Defferrard}, \bibinfo{author}{X.~Bresson},
  \bibinfo{author}{P.~Vandergheynst},
\newblock \bibinfo{title}{Convolutional neural networks on graphs with fast
  localized spectral filtering},
\newblock in: \bibinfo{booktitle}{Advances in Neural Information Processing
  Systems 29: Annual Conference on Neural Information Processing Systems 2016},
  \bibinfo{year}{2016}, pp. \bibinfo{pages}{3837--3845}.
\bibitem[{Zeng et~al.(2014)Zeng, Liu, Lai, Zhou, and Zhao}]{zeng2014relation}
\bibinfo{author}{D.~Zeng}, \bibinfo{author}{K.~Liu}, \bibinfo{author}{S.~Lai},
  \bibinfo{author}{G.~Zhou}, \bibinfo{author}{J.~Zhao},
\newblock \bibinfo{title}{Relation classification via convolutional deep neural
  network},
\newblock in: \bibinfo{booktitle}{Proceedings of the 25th International
  Conference on Computational Linguistics, {COLING} 2014},
  \bibinfo{year}{2014}, pp. \bibinfo{pages}{2335--2344}.
\bibitem[{Geng et~al.(2020)Geng, Chen, Han, Lu, and Li}]{geng2020semantic}
\bibinfo{author}{Z.~Geng}, \bibinfo{author}{G.~Chen}, \bibinfo{author}{Y.~Han},
  \bibinfo{author}{G.~Lu}, \bibinfo{author}{F.~Li},
\newblock \bibinfo{title}{Semantic relation extraction using sequential and
  tree-structured {LSTM} with attention},
\newblock \bibinfo{journal}{Inf. Sci.} \bibinfo{volume}{509}
  (\bibinfo{year}{2020}) \bibinfo{pages}{183--192}. \URLprefix
  \url{https://doi.org/10.1016/j.ins.2019.09.006}.
  \DOIprefix\doi{10.1016/j.ins.2019.09.006}.
\bibitem[{Chen et~al.(2021)Chen, Yang, Wang, Qin, Huang, and
  Zheng}]{chen2021aneur}
\bibinfo{author}{Y.~Chen}, \bibinfo{author}{W.~Yang},
  \bibinfo{author}{K.~Wang}, \bibinfo{author}{Y.~Qin},
  \bibinfo{author}{R.~Huang}, \bibinfo{author}{Q.~Zheng},
\newblock \bibinfo{title}{A neuralized feature engineering method for entity
  relation extraction},
\newblock \bibinfo{journal}{Neural Networks} \bibinfo{volume}{141}
  (\bibinfo{year}{2021}) \bibinfo{pages}{249--260}. \URLprefix
  \url{https://doi.org/10.1016/j.neunet.2021.04.010}.
  \DOIprefix\doi{10.1016/j.neunet.2021.04.010}.
\bibitem[{Xu et~al.(2021)Xu, Chen, and Zhao}]{xu2021document}
\bibinfo{author}{W.~Xu}, \bibinfo{author}{K.~Chen}, \bibinfo{author}{T.~Zhao},
\newblock \bibinfo{title}{Document-level relation extraction with
  reconstruction},
\newblock in: \bibinfo{booktitle}{Proceedings of the Thirty-Fifth {AAAI}
  Conference on Artificial Intelligence, {AAAI}}, \bibinfo{year}{2021}, pp.
  \bibinfo{pages}{14167--14175}. \URLprefix
  \url{https://ojs.aaai.org/index.php/AAAI/article/view/17667}.
\bibitem[{Huang et~al.(2021)Huang, Zhu, Feng, Ye, Lai, and
  Zhao}]{huang2021three}
\bibinfo{author}{Q.~Huang}, \bibinfo{author}{S.~Zhu},
  \bibinfo{author}{Y.~Feng}, \bibinfo{author}{Y.~Ye}, \bibinfo{author}{Y.~Lai},
  \bibinfo{author}{D.~Zhao},
\newblock \bibinfo{title}{Three sentences are all you need: Local path enhanced
  document relation extraction},
\newblock in: \bibinfo{booktitle}{Proceedings of the 59th Annual Meeting of the
  Association for Computational Linguistics and the 11th International Joint
  Conference on Natural Language Processing, {ACL/IJCNLP}},
  \bibinfo{year}{2021}, pp. \bibinfo{pages}{998--1004}. \URLprefix
  \url{https://doi.org/10.18653/v1/2021.acl-short.126}.
  \DOIprefix\doi{10.18653/v1/2021.acl-short.126}.
\bibitem[{Yang et~al.(2021)Yang, Zhang, Niu, Zhao, and Pu}]{yang2021entity}
\bibinfo{author}{S.~Yang}, \bibinfo{author}{Y.~Zhang},
  \bibinfo{author}{G.~Niu}, \bibinfo{author}{Q.~Zhao}, \bibinfo{author}{S.~Pu},
\newblock \bibinfo{title}{Entity concept-enhanced few-shot relation
  extraction},
\newblock in: \bibinfo{booktitle}{Proceedings of the 59th Annual Meeting of the
  Association for Computational Linguistics and the 11th International Joint
  Conference on Natural Language Processing, {ACL/IJCNLP}},
  \bibinfo{year}{2021}, pp. \bibinfo{pages}{987--991}. \URLprefix
  \url{https://doi.org/10.18653/v1/2021.acl-short.124}.
  \DOIprefix\doi{10.18653/v1/2021.acl-short.124}.
\bibitem[{Deng et~al.(2021)Deng, Yang, Kang, Yang, and Wu}]{Deng2021anoisy}
\bibinfo{author}{L.~Deng}, \bibinfo{author}{B.~Yang},
  \bibinfo{author}{Z.~Kang}, \bibinfo{author}{S.~Yang},
  \bibinfo{author}{S.~Wu},
\newblock \bibinfo{title}{A noisy label and negative sample robust loss
  function for dnn-based distant supervised relation extraction},
\newblock \bibinfo{journal}{Neural Networks} \bibinfo{volume}{139}
  (\bibinfo{year}{2021}) \bibinfo{pages}{358--370}. \URLprefix
  \url{https://doi.org/10.1016/j.neunet.2021.03.030}.
  \DOIprefix\doi{10.1016/j.neunet.2021.03.030}.
\bibitem[{Zhou et~al.(2021)Zhou, Pan, Bai, Luo, and Wu}]{zhou2021selfselect}
\bibinfo{author}{Y.~Zhou}, \bibinfo{author}{L.~Pan}, \bibinfo{author}{C.~Bai},
  \bibinfo{author}{S.~Luo}, \bibinfo{author}{Z.~Wu},
\newblock \bibinfo{title}{Self-selective attention using correlation between
  instances for distant supervision relation extraction},
\newblock \bibinfo{journal}{Neural Networks} \bibinfo{volume}{142}
  (\bibinfo{year}{2021}) \bibinfo{pages}{213--220}. \URLprefix
  \url{https://doi.org/10.1016/j.neunet.2021.04.032}.
  \DOIprefix\doi{10.1016/j.neunet.2021.04.032}.
\bibitem[{Surdeanu et~al.(2012)Surdeanu, Tibshirani, Nallapati, and
  Manning}]{Surdeanu2012multi}
\bibinfo{author}{M.~Surdeanu}, \bibinfo{author}{J.~Tibshirani},
  \bibinfo{author}{R.~Nallapati}, \bibinfo{author}{C.~D. Manning},
\newblock \bibinfo{title}{Multi-instance multi-label learning for relation
  extraction},
\newblock in: \bibinfo{booktitle}{Proceedings of the 2012 Joint Conference on
  Empirical Methods in Natural Language Processing and Computational Natural
  Language Learning, {EMNLP-CoNLL} 2012}, \bibinfo{publisher}{{ACL}},
  \bibinfo{year}{2012}, pp. \bibinfo{pages}{455--465}.
\bibitem[{Zeng et~al.(2015)Zeng, Liu, Chen, and Zhao}]{zeng2015distant}
\bibinfo{author}{D.~Zeng}, \bibinfo{author}{K.~Liu}, \bibinfo{author}{Y.~Chen},
  \bibinfo{author}{J.~Zhao},
\newblock \bibinfo{title}{Distant supervision for relation extraction via
  piecewise convolutional neural networks},
\newblock in: \bibinfo{booktitle}{Proceedings of the 2015 Conference on
  Empirical Methods in Natural Language Processing, {EMNLP} 2015},
  \bibinfo{publisher}{The Association for Computational Linguistics},
  \bibinfo{year}{2015}, pp. \bibinfo{pages}{1753--1762}.
  \DOIprefix\doi{10.18653/v1/d15-1203}.
\bibitem[{Yuan et~al.(2019)Yuan, Liu, Tang, Zhang, Zhuang, Pu, Wu, and
  Ren}]{yuan2019c2sa}
\bibinfo{author}{Y.~Yuan}, \bibinfo{author}{L.~Liu}, \bibinfo{author}{S.~Tang},
  \bibinfo{author}{Z.~Zhang}, \bibinfo{author}{Y.~Zhuang},
  \bibinfo{author}{S.~Pu}, \bibinfo{author}{F.~Wu}, \bibinfo{author}{X.~Ren},
\newblock \bibinfo{title}{Cross-relation cross-bag attention for
  distantly-supervised relation extraction},
\newblock in: \bibinfo{booktitle}{Proceedings of the Thirty-Third {AAAI}
  Conference on Artificial Intelligence, {AAAI} 2019},
  \bibinfo{publisher}{{AAAI} Press}, \bibinfo{year}{2019}, pp.
  \bibinfo{pages}{419--426}. \DOIprefix\doi{10.1609/aaai.v33i01.3301419}.
\bibitem[{Ye and Ling(2019)}]{ye2019distant}
\bibinfo{author}{Z.~Ye}, \bibinfo{author}{Z.~Ling},
\newblock \bibinfo{title}{Distant supervision relation extraction with
  intra-bag and inter-bag attentions},
\newblock in: \bibinfo{booktitle}{Proceedings of the 2019 Conference of the
  North American Chapter of the Association for Computational Linguistics:
  Human Language Technologies, {NAACL-HLT} 2019},
  \bibinfo{publisher}{Association for Computational Linguistics},
  \bibinfo{year}{2019}, pp. \bibinfo{pages}{2810--2819}.
  \DOIprefix\doi{10.18653/v1/n19-1288}.
\bibitem[{Dai et~al.(2019)Dai, Xu, and Song}]{dai2019feature}
\bibinfo{author}{L.~Dai}, \bibinfo{author}{B.~Xu}, \bibinfo{author}{H.~Song},
\newblock \bibinfo{title}{Feature-level attention based sentence encoding for
  neural relation extraction},
\newblock in: \bibinfo{booktitle}{Natural Language Processing and Chinese
  Computing - 8th {CCF} International Conference, {NLPCC} 2019}, volume
  \bibinfo{volume}{11838} of \textit{\bibinfo{series}{Lecture Notes in Computer
  Science}}, \bibinfo{publisher}{Springer}, \bibinfo{year}{2019}, pp.
  \bibinfo{pages}{184--196}. \DOIprefix\doi{10.1007/978-3-030-32233-5\_15}.
\bibitem[{Yu et~al.(2019)Yu, Zhang, Liu, Wang, Li, and Li}]{yu2019beyond}
\bibinfo{author}{B.~Yu}, \bibinfo{author}{Z.~Zhang}, \bibinfo{author}{T.~Liu},
  \bibinfo{author}{B.~Wang}, \bibinfo{author}{S.~Li}, \bibinfo{author}{Q.~Li},
\newblock \bibinfo{title}{Beyond word attention: Using segment attention in
  neural relation extraction},
\newblock in: \bibinfo{booktitle}{Proceedings of the Twenty-Eighth
  International Joint Conference on Artificial Intelligence, {IJCAI} 2019},
  \bibinfo{publisher}{ijcai.org}, \bibinfo{year}{2019}, pp.
  \bibinfo{pages}{5401--5407}. \DOIprefix\doi{10.24963/ijcai.2019/750}.
\bibitem[{Peng et~al.(2022)Peng, Han, Cui, Yue, Han, and Liu}]{peng2022ghe-lpc}
\bibinfo{author}{T.~Peng}, \bibinfo{author}{R.~Han}, \bibinfo{author}{H.~Cui},
  \bibinfo{author}{L.~Yue}, \bibinfo{author}{J.~Han}, \bibinfo{author}{L.~Liu},
\newblock \bibinfo{title}{Distantly supervised relation extraction using global
  hierarchy embeddings and local probability constraints},
\newblock \bibinfo{journal}{Knowl. Based Syst.} \bibinfo{volume}{235}
  (\bibinfo{year}{2022}) \bibinfo{pages}{107637}. \URLprefix
  \url{https://doi.org/10.1016/j.knosys.2021.107637}.
  \DOIprefix\doi{10.1016/j.knosys.2021.107637}.
\bibitem[{Mikolov et~al.(2013)Mikolov, Sutskever, Chen, Corrado, and
  Dean}]{mikolov2013distributed}
\bibinfo{author}{T.~Mikolov}, \bibinfo{author}{I.~Sutskever},
  \bibinfo{author}{K.~Chen}, \bibinfo{author}{G.~S. Corrado},
  \bibinfo{author}{J.~Dean},
\newblock \bibinfo{title}{Distributed representations of words and phrases and
  their compositionality},
\newblock in: \bibinfo{booktitle}{Proceedings of the 27th Annual Conference on
  Neural Information Processing Systems , {NIPS} 2013}, \bibinfo{year}{2013},
  pp. \bibinfo{pages}{3111--3119}.
\bibitem[{Li et~al.(2020)Li, Long, Shen, Zhou, Yao, Huo, and Jiang}]{li2020seg}
\bibinfo{author}{Y.~Li}, \bibinfo{author}{G.~Long}, \bibinfo{author}{T.~Shen},
  \bibinfo{author}{T.~Zhou}, \bibinfo{author}{L.~Yao},
  \bibinfo{author}{H.~Huo}, \bibinfo{author}{J.~Jiang},
\newblock \bibinfo{title}{Self-attention enhanced selective gate with
  entity-aware embedding for distantly supervised relation extraction},
\newblock in: \bibinfo{booktitle}{Proceedings of the Thirty-Fourth {AAAI}
  Conference on Artificial Intelligence, {AAAI} 2020},
  \bibinfo{publisher}{{AAAI} Press}, \bibinfo{year}{2020}, pp.
  \bibinfo{pages}{8269--8276}.
\bibitem[{He et~al.(2016)He, Zhang, Ren, and Sun}]{he2016residual}
\bibinfo{author}{K.~He}, \bibinfo{author}{X.~Zhang}, \bibinfo{author}{S.~Ren},
  \bibinfo{author}{J.~Sun},
\newblock \bibinfo{title}{Deep residual learning for image recognition},
\newblock in: \bibinfo{booktitle}{Proceedings of the 2016 {IEEE} Conference on
  Computer Vision and Pattern Recognition, {CVPR} 2016},
  \bibinfo{publisher}{{IEEE} Computer Society}, \bibinfo{year}{2016}, pp.
  \bibinfo{pages}{770--778}. \DOIprefix\doi{10.1109/CVPR.2016.90}.
\bibitem[{Ba et~al.(2016)Ba, Kiros, and Hinton}]{ba2016layer}
\bibinfo{author}{L.~J. Ba}, \bibinfo{author}{J.~R. Kiros},
  \bibinfo{author}{G.~E. Hinton},
\newblock \bibinfo{title}{Layer normalization},
\newblock \bibinfo{journal}{CoRR} \bibinfo{volume}{abs/1607.06450}
  (\bibinfo{year}{2016}). \href{http://arxiv.org/abs/1607.06450}{{\tt
  arXiv:1607.06450}}.
\bibitem[{Lin et~al.(2017)Lin, Feng, dos Santos, Yu, Xiang, Zhou, and
  Bengio}]{lin2017self}
\bibinfo{author}{Z.~Lin}, \bibinfo{author}{M.~Feng}, \bibinfo{author}{C.~N. dos
  Santos}, \bibinfo{author}{M.~Yu}, \bibinfo{author}{B.~Xiang},
  \bibinfo{author}{B.~Zhou}, \bibinfo{author}{Y.~Bengio},
\newblock \bibinfo{title}{A structured self-attentive sentence embedding},
\newblock in: \bibinfo{booktitle}{Proceedings of the 5th International
  Conference on Learning Representations, {ICLR} 2017},
  \bibinfo{publisher}{OpenReview.net}, \bibinfo{year}{2017}.
\bibitem[{Shen et~al.(2018)Shen, Zhou, Long, Jiang, Pan, and
  Zhang}]{shen2018disan}
\bibinfo{author}{T.~Shen}, \bibinfo{author}{T.~Zhou},
  \bibinfo{author}{G.~Long}, \bibinfo{author}{J.~Jiang},
  \bibinfo{author}{S.~Pan}, \bibinfo{author}{C.~Zhang},
\newblock \bibinfo{title}{Disan: Directional self-attention network for
  rnn/cnn-free language understanding},
\newblock in: \bibinfo{booktitle}{Proceedings of the Thirty-Second {AAAI}
  Conference on Artificial Intelligence, {AAAI} 2018},
  \bibinfo{publisher}{{AAAI} Press}, \bibinfo{year}{2018}, pp.
  \bibinfo{pages}{5446--5455}.
\bibitem[{Jat et~al.(2018)Jat, Khandelwal, and Talukdar}]{jat2018improving}
\bibinfo{author}{S.~Jat}, \bibinfo{author}{S.~Khandelwal},
  \bibinfo{author}{P.~P. Talukdar},
\newblock \bibinfo{title}{Improving distantly supervised relation extraction
  using word and entity based attention},
\newblock \bibinfo{journal}{CoRR} \bibinfo{volume}{abs/1804.06987}
  (\bibinfo{year}{2018}). \href{http://arxiv.org/abs/1804.06987}{{\tt
  arXiv:1804.06987}}.
\bibitem[{Zhu et~al.(2020)Zhu, Wang, Yu, Zhou, Chen, Zhang, and
  Zhang}]{zhu2020towards}
\bibinfo{author}{T.~Zhu}, \bibinfo{author}{H.~Wang}, \bibinfo{author}{J.~Yu},
  \bibinfo{author}{X.~Zhou}, \bibinfo{author}{W.~Chen},
  \bibinfo{author}{W.~Zhang}, \bibinfo{author}{M.~Zhang},
\newblock \bibinfo{title}{Towards accurate and consistent evaluation: {A}
  dataset for distantly-supervised relation extraction},
\newblock in: \bibinfo{booktitle}{Proceedings of the 28th International
  Conference on Computational Linguistics, {COLING} 2020},
  \bibinfo{publisher}{International Committee on Computational Linguistics},
  \bibinfo{year}{2020}, pp. \bibinfo{pages}{6436--6447}.
  \DOIprefix\doi{10.18653/v1/2020.coling-main.566}.
\bibitem[{Yao et~al.(2019)Yao, Ye, Li, Han, Lin, Liu, Liu, Huang, Zhou, and
  Sun}]{yao2019docred}
\bibinfo{author}{Y.~Yao}, \bibinfo{author}{D.~Ye}, \bibinfo{author}{P.~Li},
  \bibinfo{author}{X.~Han}, \bibinfo{author}{Y.~Lin}, \bibinfo{author}{Z.~Liu},
  \bibinfo{author}{Z.~Liu}, \bibinfo{author}{L.~Huang},
  \bibinfo{author}{J.~Zhou}, \bibinfo{author}{M.~Sun},
\newblock \bibinfo{title}{Docred: {A} large-scale document-level relation
  extraction dataset},
\newblock in: \bibinfo{booktitle}{Proceedings of the 57th Conference of the
  Association for Computational Linguistics, {ACL}}, \bibinfo{year}{2019}, pp.
  \bibinfo{pages}{764--777}. \URLprefix
  \url{https://doi.org/10.18653/v1/p19-1074}.
  \DOIprefix\doi{10.18653/v1/p19-1074}.
\bibitem[{Srivastava et~al.(2014)Srivastava, Hinton, Krizhevsky, Sutskever, and
  Salakhutdinov}]{srivastava2014dropout}
\bibinfo{author}{N.~Srivastava}, \bibinfo{author}{G.~E. Hinton},
  \bibinfo{author}{A.~Krizhevsky}, \bibinfo{author}{I.~Sutskever},
  \bibinfo{author}{R.~Salakhutdinov},
\newblock \bibinfo{title}{Dropout: a simple way to prevent neural networks from
  overfitting},
\newblock \bibinfo{journal}{J. Mach. Learn. Res.} \bibinfo{volume}{15}
  (\bibinfo{year}{2014}) \bibinfo{pages}{1929--1958}.
\bibitem[{Cotter et~al.(2011)Cotter, Shamir, Srebro, and
  Sridharan}]{cotter2011minibatch}
\bibinfo{author}{A.~Cotter}, \bibinfo{author}{O.~Shamir},
  \bibinfo{author}{N.~Srebro}, \bibinfo{author}{K.~Sridharan},
\newblock \bibinfo{title}{Better mini-batch algorithms via accelerated gradient
  methods},
\newblock in: \bibinfo{booktitle}{Proceedings of the 25th Annual Conference on
  Neural Information Processing Systems, {NIPS} 2011}, \bibinfo{year}{2011},
  pp. \bibinfo{pages}{1647--1655}.

\end{thebibliography}

\end{document}